\documentclass[runningheads]{llncs}
\usepackage{graphicx}
\usepackage{comment}
\usepackage{epsfig}
\usepackage{amsmath,amssymb} % define this before the line numbering.
\usepackage{color}
\usepackage{multirow}
\usepackage{booktabs}
\usepackage{subfigure}
\usepackage{url}
\def\vs{\emph{vs.\ }}
\def\eg{\emph{e.g.\ }}
\def\ie{\emph{i.e.\ }}

\def\etal{\emph{et al.\ }}

\begin{document}
% \linenumbers
\pagestyle{headings}
\mainmatter
\def\ECCVSubNumber{637}  % Insert your submission number here

\title{DHP: Differentiable Meta Pruning via HyperNetworks} % Replace with your title

%\begin{comment}

\titlerunning{DHP: Differentiable Meta Pruning via HyperNetworks} 
\authorrunning{Yawei Li, Shuhang Gu, Kai Zhang, Luc Van Gool, Radu Timofte} 
\author{Yawei Li$^{1}$, Shuhang Gu$^{1,2}$, Kai Zhang$^1$, Luc Van Gool$^{1,3}$, Radu Timofte$^1$}
\institute{$^1$Computer Vision Lab, ETH Z\"urich, $^2$The University of Sydney, $^3$KU Leuven \\
\email{\{yawei.li, kai.zhang, vangool, radu.timofte\}@vision.ee.ethz.ch} \\ \email{shuhanggu@gmail.com}}

%\end{comment}
%******************

\begin{comment}
\titlerunning{Abbreviated paper title}
\author{First Author\inst{1}\orcidID{0000-1111-2222-3333} \and
Second Author\inst{2,3}\orcidID{1111-2222-3333-4444} \and
Third Author\inst{3}\orcidID{2222--3333-4444-5555}}
%
\authorrunning{F. Author et al.}
\institute{Princeton University, Princeton NJ 08544, USA \and
Springer Heidelberg, Tiergartenstr. 17, 69121 Heidelberg, Germany
\email{lncs@springer.com}\\
\url{http://www.springer.com/gp/computer-science/lncs} \and
ABC Institute, Rupert-Karls-University Heidelberg, Heidelberg, Germany\\
\email{\{abc,lncs\}@uni-heidelberg.de}}
\end{comment}
%******************
\maketitle

\begin{abstract}
Network pruning has been the driving force for the acceleration of neural networks and the alleviation of model storage/transmission burden. With the advent of AutoML and neural architecture search (NAS), pruning has become topical with automatic mechanism and searching based architecture optimization. Yet, current automatic designs rely on either reinforcement learning or evolutionary algorithm. Due to the non-differentiability of those algorithms, the pruning algorithm needs a long searching stage before reaching the convergence.

To circumvent this problem, this paper introduces a differentiable pruning method via hypernetworks for automatic network pruning. The specifically designed hypernetworks take latent vectors as input and generate the weight parameters of the backbone network. The latent vectors control the output channels of the convolutional layers in the backbone network and act as a handle for the pruning of the layers. By enforcing $\ell_1$ sparsity regularization to the latent vectors and utilizing proximal gradient solver, sparse latent vectors can be obtained. Passing the sparsified latent vectors through the hypernetworks, the corresponding slices of the generated weight parameters can be removed, achieving the effect of network pruning. The latent vectors of all the layers are pruned together, resulting in an automatic layer configuration. Extensive experiments are conducted on various networks for image classification, single image super-resolution, and denoising. And the experimental results validate the proposed method. Code is available at \textcolor{red}{\url{https://github.com/ofsoundof/dhp}}.

\end{abstract}

\section{Introduction}
\label{sec:introduction}
\begin{figure}
\centering     %%% not \center
\subfigure[FLOPs compression ratio.]{\label{fig:mobilenet_dhp_flops}\includegraphics[width=0.4\textwidth]{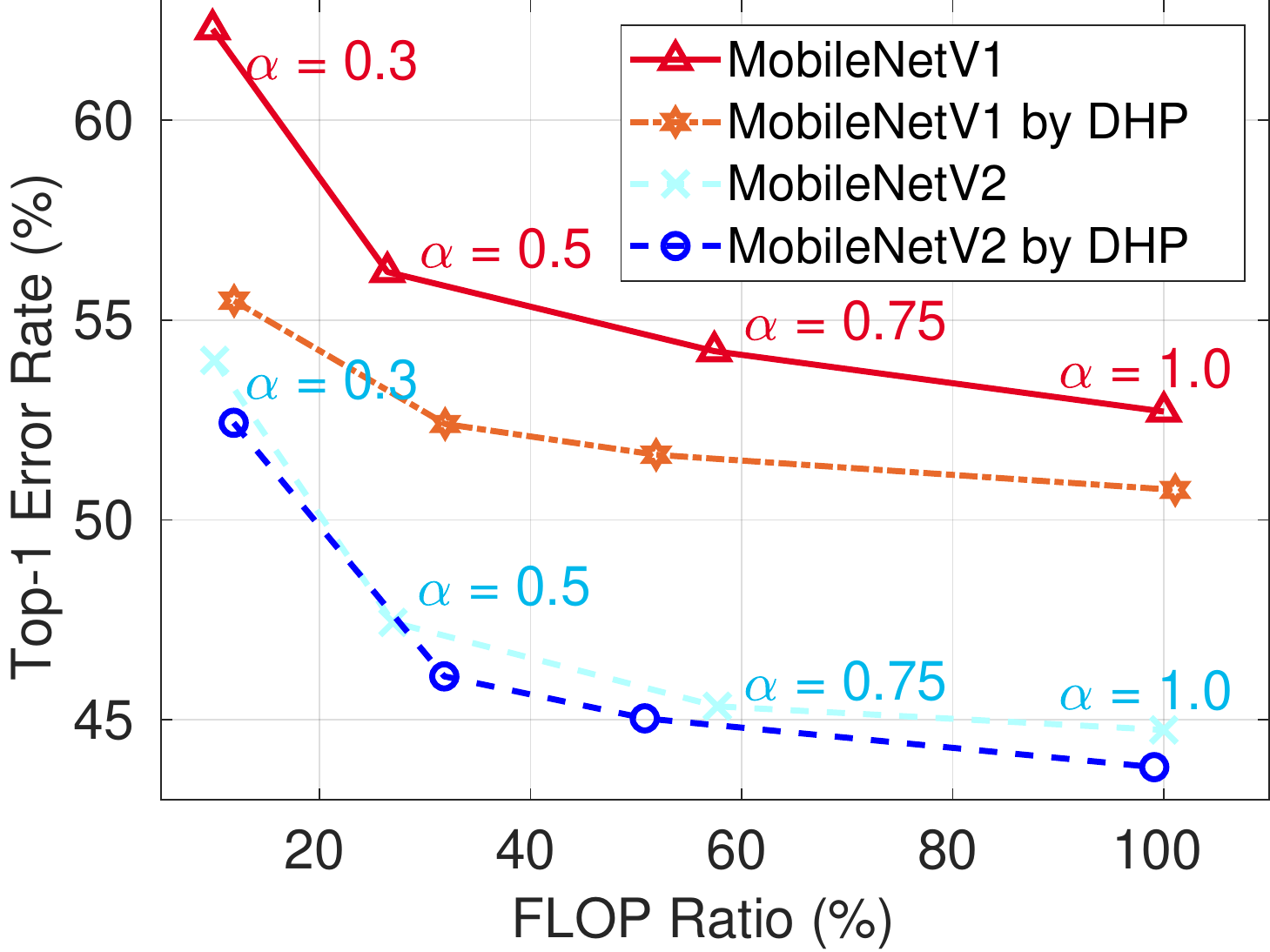}}
\subfigure[Parameter compression ratio.]{\label{fig:mobilenet_dhp_parameters}\includegraphics[width=0.4\textwidth]{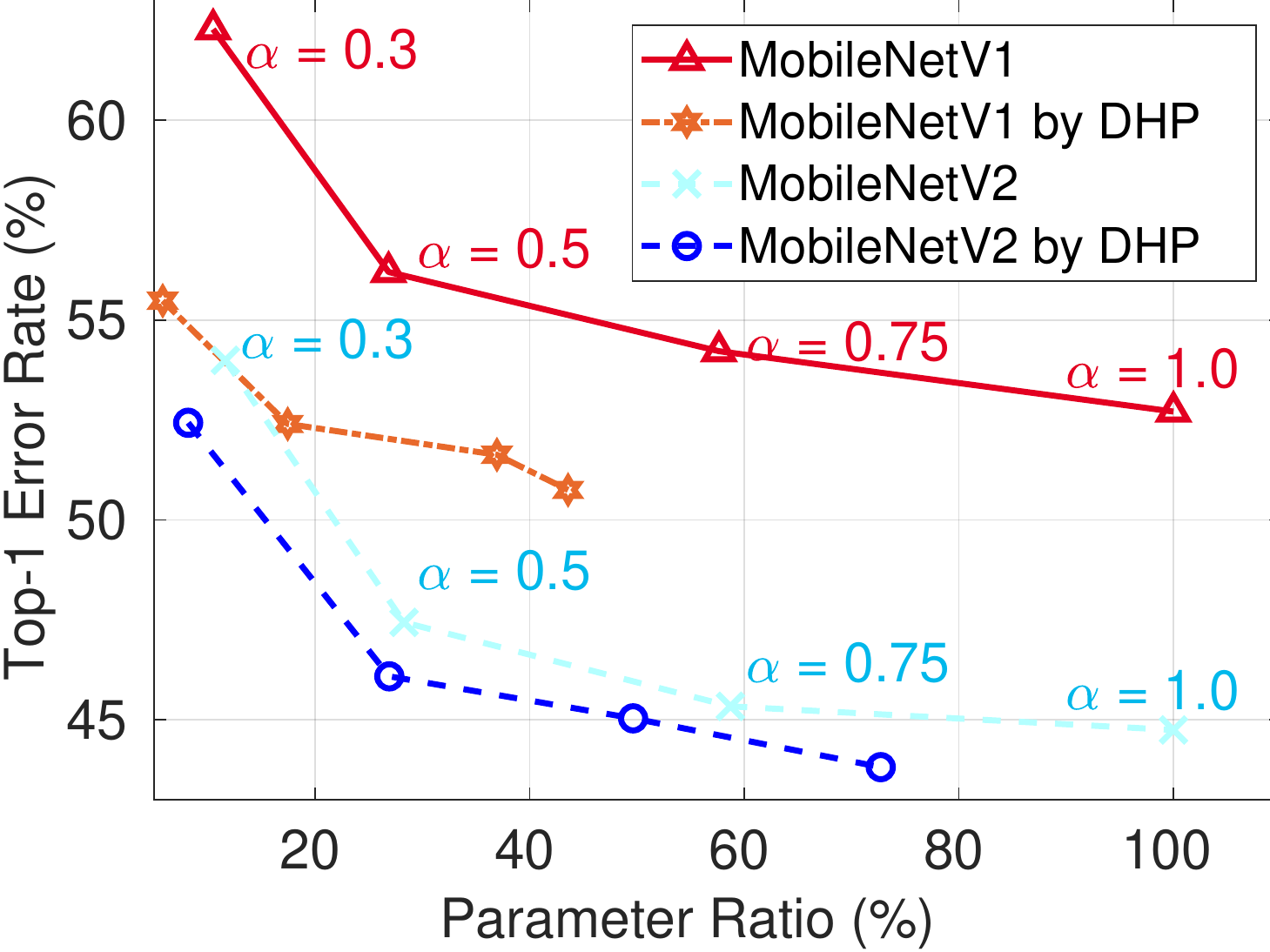}}
% \vspace{-0.2cm}
\caption{Top-1 error \vs FLOPs and parameter compression ratio on MobileNets. The original model with different width multipliers $\alpha$ is set as the baseline. The DHP operating points near 100\% FLOPs ratio is obtained by pruning the two networks with $\alpha = 2$. The DHP models outperforms the original at all the operating points}
% \vspace{-0.4cm}
\label{fig:mobilenet_dhp_comparison}
\end{figure}
These days, network pruning has become the workhorse for network compression, which aims at lightweight and efficient model for fast inference~\cite{han2015deep,he2017channel,liu2019rethinking,li2020group}. 
This is of particular importance for the deployment of tiny artificial intelligence (Tiny AI) algorithms on smart phones and edge devices~\cite{mitreview2020}. 
Since the emerging of network pruning a couple of methods have been proposed based on the analysis of gradients, Hessians or filter distribution~\cite{lecun1990optimal,hassibi1993second,dong2017learning,molchanov2019importance,he2019filter}. 
With the advent of AutoML and neural architecture search (NAS)~\cite{zoph2017nas,chen2018constraint}, a new trend of network pruning emerges, \ie pruning with automatic algorithms and targeting distinguishing sub-architectures.
Among them, reinforcement learning and evolutionary algorithm become the natural choice~\cite{He_2018_ECCV_AMC,liu2019metapruning}. 
The core idea is to search a certain fine-grained layer-wise distinguishing configuration among the all of the possible choices (population in the terminology of evolutionary algorithm). 
After the searching stage, the candidate that optimizes the network prediction accuracy under constrained budgets is chosen.

The advantage of these automatic pruning methods is the final layer-wise distinguishing configuration. 
Thus, hand-crafted design is no longer necessary. 
However, the main concern of these algorithms is the convergence property. 
For example, reinforcement learning is notorious for its difficulty of convergence under large or even middle level number of states~\cite{sutton2018reinforcement}. 
Evolutionary algorithm needs to choose the best candidate from the already converged algorithm. 
But the dilemma lies in the impossibility of training the whole population till convergence and the difficulty of choosing the best candidate from unconverged population~\cite{liu2019metapruning,hayashi2019einconv}. 
A promising solution to this problem is endowing the searching mechanism with differentiability or resorting to an approximately differentiable algorithm. 
This is due to the fact that differentiability has the potential to make the searching stage efficient. 
Actually, differentiability has facilitated a couple of machine learning approaches and the typical one among them is NAS. 
Early works on NAS have insatiable demand for computing resources, consuming tens of thousands of GPU hours for a satisfactory convergence~\cite{zoph2017nas,zoph2018nas}. 
Differentiable architecture search (DARTS) reduces the insatiable consumption to tens of GPU hours, which has boosted the development of NAS during the past year~\cite{liu2019darts}.

Another noteworthy direction for automatic pruning is brought by MetaPruning~\cite{liu2019metapruning} which introduces hypernetworks~\cite{ha2017hypernetworks} into network compression. 
The output of the so-called hypernetwork is used as the parameters of the backbone network. 
During training, the gradients are also back-propagated to the hypernetworks. 
This method falls in the paradigm of meta learning since the parameters in the hypernetwork act as the meta-data of the parameters in the backbone network. 
But the problem of this method is that the hypernetworks can only output fixed-size weights, which cannot serve as a layer-wise configuration searching mechanism. 
Thus, a searching algorithm such as evolutionary algorithm is necessary for the discovery of a good candidate. 
Although this is quite a natural choice, there is still one interesting question, namely, whether one can design a hypernetwork whose output size depends on the input (termed as latent vector in this paper) so that by only dealing with the latent vector, the backbone network can be automatically pruned.
\begin{figure}[!t]
\centering
\includegraphics[width=0.8\linewidth]{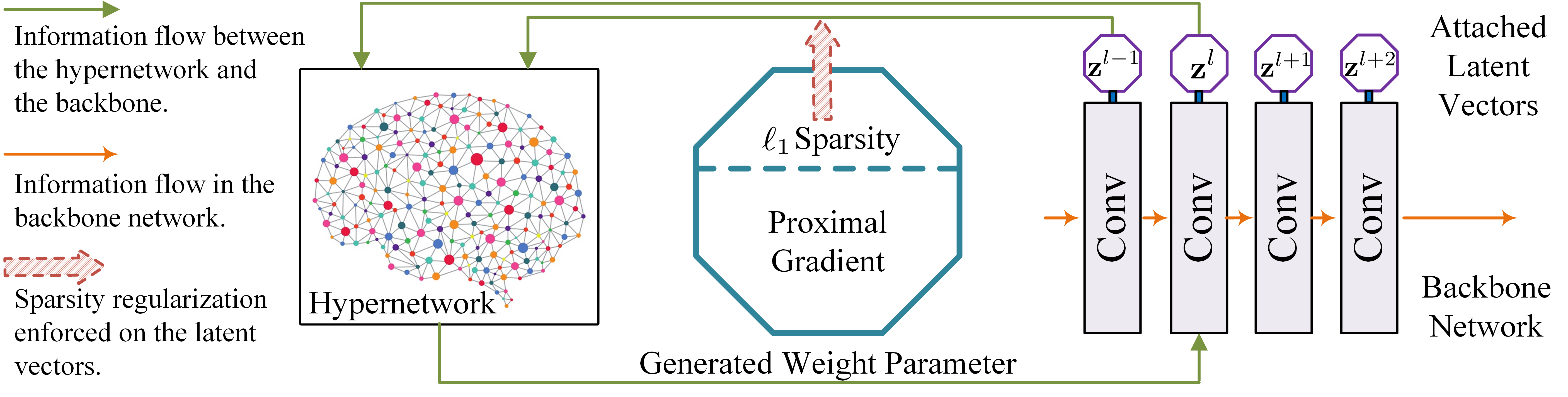}
% \vspace{-0.2cm}
\caption{The workflow of the proposed differentiable pruning method. The latent vectors $\mathbf{z}$ attached to the convolutional layers act as the handle for network pruning. The hypernetwork takes two latent vectors as input and emits output as the weight of the backbone layer. $\ell_1$ sparsity regularization is enforced on the latent vectors. The differentiability comes with the hypernetwork tailored to pruning and the proximal gradient exploited to solve problem. After the pruning stage, sparse latent vectors are obtained which result in pruned weights after being passed through the hypernetwork}
% \vspace{-0.4cm}
\label{fig:dhp_workflow}
\end{figure}

To solve the aforementioned problem, we propose the differentiable meta pruning approach via hypernetworks (DHP, D -- \textbf{D}ifferentiable, H -- \textbf{H}yper, P -- \textbf{P}runing) shown in Fig.~\ref{fig:dhp_workflow}. 
A new design of hypernetwork is proposed to adapt to the requirements of differentiability. 
Each layer is endowed with a latent vector that controls the output channels of this layer. 
Since the layers in the network are connected, the latent vector also controls the input channel of the next layer.
The hypernetwork takes as input the latent vectors of the current layer and previous layer that controls the output and input channels of the current layer respectively. 
Passing the latent vectors through the hypernetwork leads to outputs which are used as the parameters of the backbone network. 
To achieve the effect of automatic pruning, $\ell_1$ sparsity regularizer is applied to the latent vectors. 
A pruned model is discovered by updating the latent vectors with proximal gradient.
The searching stage stops when the compression ratio drops to the target level. 
After the searching stage, the latent vectors are sparsified.
Accordingly, the outputs of the hypernetworks that are covariant with the latent vector are also compressed.
The advantage of the proposed method is that it is only necessary to deal with the latent vectors, which automates network pruning without the other bells and whistles.

With the fast development of efficient network design and NAS, the usefulness of network pruning is frequently challenged. 
But by analyzing the performance on MobileNetV1~\cite{howard2017mobilenets} and MobileNetV2~\cite{sandler2018mobilenetv2} in Fig.~\ref{fig:mobilenet_dhp_comparison}, we conclude that automatic network pruning is of vital importance for further exploring the capacity of efficient networks.
Efficient network design and NAS can only result in an overall architecture with building blocks endowed with the same sub-architecture. 
By automatic network pruning, the efficient networks obtained by either human experts or NAS can be further compressed, leading to layer-wise distinguishing configurations, which can be seen as a fine-grained architecture search.

Thus, the contribution of this paper is as follows.
\begin{itemize}
%   \vspace{-0.2cm}
  \item[I] A new architecture of hypernetwork is designed. Different from the classical hypernetwork composed of linear layers, the new design is tailored to automatic network pruning. By only operating on the input of the hypernetwork, the backbone network can be pruned. 
  
  \item[II] A differentiable automatic networking pruning method is proposed. The differentiability comes with the designed hypernetwork and the utilized proximal gradient. It accelerates the convergence of the pruning algorithm.
  
  \item[III] By the experiments on various vision tasks and modern convolutional neural networks (CNNs)~\cite{he2016deep,huang2017densely,howard2017mobilenets,sandler2018mobilenetv2,zhang2017beyond,ronneberger2015unet,ledig2016photo,lim2017enhanced}, the potential of automatic network pruning as fine-grained architecture search is revealed.
%   \vspace{-0.2cm}
\end{itemize}

\section{Related Works}
\label{sec:related_works}

\textbf{Network pruning.}
Aiming at removing the weak filter connections that have the least influence on the accuracy of the network, network pruning attracts increasing attention.
Early attempts emphasize more on the storage consumption, various criteria have been explored to remove inconsequential connections in an unstructural manner~\cite{han2015deep,liu2015sparse}.
Despite their success in reducing network parameters, unstructural pruning leads to irregular weight parameters, limited in the actual acceleration of the pruned network.
To further address the efficiency issue, structured pruning methods directly zero out structured groups of the convolutional filters.
For example, Wen~\etal~\cite{wen2016learning} and Alvarez~\etal~\cite{alvarez2016learning} firstly proposed to resort to group sparsity regularization during training to reduce the number of feature maps in each layer.
Since that, the field has witnessed a variety of regularization strategies\cite{alvarez2017compression,yoon2017combined,li2019oicsr,torfi2019gasl,li2020group}.
These elaborately designed regularization methods considerably advance the pruning performance.
But they often rely on carefully adjusted hyper-parameters selected for specific network architecture and dataset.

\textbf{AutoML.} Recently, there is an emerging trend of exploiting AutoML for automatic network compression~\cite{He_2018_ECCV_AMC,liu2019metapruning,hayashi2019einconv,chin2020towards}. 
The rationality lies in the exploration among the total population of network configurations for a final best candidate.
He~\etal exploited reinforcement learning agents to prune the networks where hand-crafted design is not longer necessary~\cite{He_2018_ECCV_AMC}. 
Hayashi~\etal utilized genetic algorithm to enumerate candidate in the designed hypergraph for tensor network decomposition~\cite{hayashi2019einconv}. 
Liu~\etal trained a hypernetwork to generate weights of the backbone network and used evolutionary algorithm to search for the best candidate~\cite{liu2019metapruning}. 
The problem of these approachs is that the searching algorithms are not differentiable, which does not result in guaranteed convergence.

\textbf{NAS.}
NAS automatizes the manual task of neural network architecture design. 
Optimally, searched networks achieve smaller test error, require fewer parameters and need less computations than their manually designed counterparts~\cite{zoph2017nas,real2019regularized}.
But the main drawback of early strategies is their almost insatiable demand for computational resources.
To alleviate the computational burden several methods~\cite{zoph2018nas,liu2018nas,liu2019darts} are proposed to search for a basic building block, \ie cell, opposed to an entire network.
Then, stacking multiple cells with equivalent structure but different weights defines a full network~\cite{pham2018efficient,brock2018smash}.
Another recent trend in NAS is differentiable search methods such as DARTS~\cite{liu2019darts}. The differentiability allows the fast convergence of the searching algorithm and thus boosts the fast development of NAS during the past year. In this paper we propose a differentiable counterpart for automatic network pruning.

\textbf{Meta learning and hypernetworks.}
Meta learning is a broad family of machine learning techniques that deal with the problem of learning to learn. 
An emerging trend of meta learning uses hypernetworks to predict the weight parameters in the backbone network~\cite{ha2017hypernetworks}. 
Since the introduction of hypernetworks, it has found wide applications in NAS~\cite{brock2018smash}, multi-task learning~\cite{pan2018hyperst}, Bayesian neural networks~\cite{krueger2017bayesian}, and also network pruning~\cite{liu2019metapruning}. 
In this paper, we propose a new design of hypernetwork which is especially suitable for network pruning and makes differentiability possible for automatic network pruning.   

\section{Methodology}
\label{sec:methodology}

The pipeline of the proposed method is shown in Fig~\ref{fig:dhp_workflow}. 
The two cores of the whole pipeline are the designed hypernetwork and the optimization algorithm. 
In the forward pass, the designed hypernetwork takes as input the latent vectors and predicts the weight parameters for the backbone network. 
In the backward pass, the gradients are back-propagated to the hypernetwork. 
The $\ell_1$ sparsity regularizer is enforced on the latent vectors and proximal gradient is used to solve the problem. 
The dimension of the output of the hypernetwork is covariant with that of the input. 
Due to this property, the output weights are pruned along with the sparsified latent vectors after the optimization step. 
The differentiability comes with the covariance property of the hypernetworks, the $\ell_1$ sparsity regularization enforced on the latent vectors, and the proximal gradient used to solve the problem. 
The automation of pruning is due to the fact that all of the latent vectors are non-discriminatively regularized and that proximal gradient discovers the potential less important elements automatically.

\subsection{Hypernetwork design}
\label{subsec:hypernetwork}

\begin{figure}[!t]
% \vspace{-0.2cm}
\centering
\includegraphics[width=0.8\linewidth]{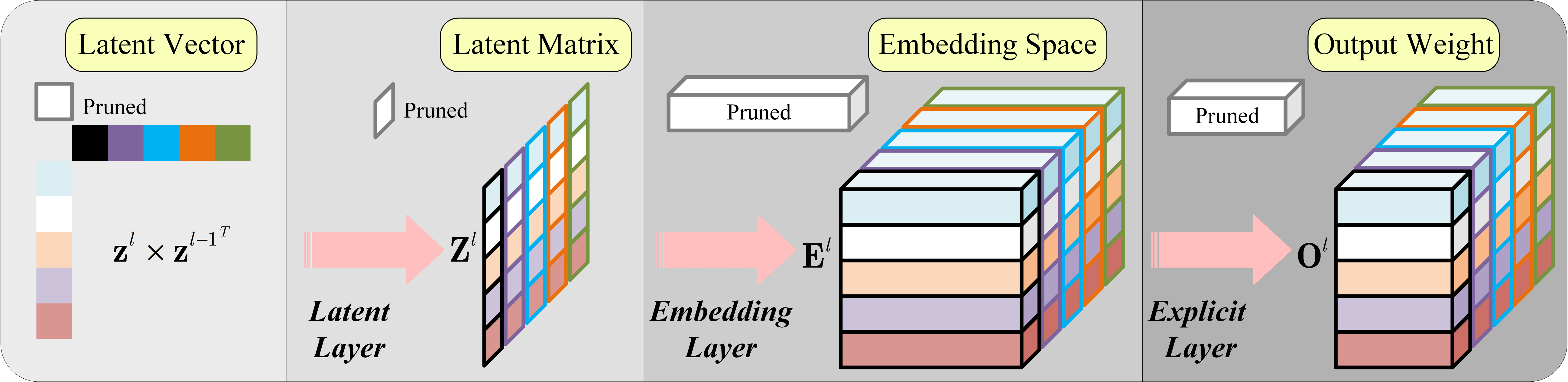}
% \vspace{-0.2cm}
\caption{Illustration of the hypernetwork designed for network pruning. It generates a weight tensor after passing the input latent vector through the latent layer, the embedding layer, and the explicit layer. If one element in $\mathbf{z}^l$ is pruned, the corresponding slice of the output tensor is also pruned. See Subsec.~\ref{subsec:hypernetwork} for details}
% \vspace{-0.4cm}
\label{fig:hypernetwork_design}
\end{figure}

\textbf{Notation:} 
Unless otherwise stated, we use the normal ($x$), minuscule bold ($\mathbf{z}$), and capital bold ($\mathbf{Z}$) letters to denote scalars, vectors, and matrices/high-dimensional tensors. 
The elements of a matrix/tensor is indexed by the subscript as $\mathbf{Z}_{i,j}$ which could be scalars or vectors depending on the the dimension of the indexed subject.

We first introduce the design of the hypernetwork shown in Fig.~\ref{fig:hypernetwork_design}. 
In summary, the hypernetwork consists of three layers. 
The latent layer takes as input the latent vectors and computes a latent matrix from them. 
The embedding layer projects the elements of the latent matrix to an embedding space. 
The last explicit layer converts the embedded vectors to the final output.
This design is inspired by fully connected layers in~\cite{ha2017hypernetworks,liu2019metapruning} but differs from those designs in that the output dimension is covariant with the input latent vector. 
This design is applicable to all types of convolutions including the standard convolution, depth-wise convolution, point-wise convolution, and transposed convolution. 
% 
% And for the simplicity of reference, the term convolution is used to denote any of them. 

Suppose that the given is an $L$-layer CNN. 
The dimension of the weight parameter of the $l$-th convolutional layer is $n \times c \times w \times h$, where $n$, $c$, and $w \times h$ denote the output channel, input channel, and kernel size of the layer, respectively. 
Every layer is endowed with a latent vector $\mathbf{z}^l$. 
The latent vector has the same size as the output channel of the layer, \ie, $\mathbf{z}^l \in \mathbb{R}^n$. 
Thus, the previous layer is given a latent vector $\mathbf{z}^{l-1} \in \mathbb{R}^c$. 
The hypernetwork receives the latent vectors $\mathbf{z}^l$ and $\mathbf{z}^{l-1}$ of the current and the previous layer as input.
A latent matrix is first computed from the two latent vectors, namely,
\begin{equation}
    \mathbf{Z}^l = {\mathbf{z}^{l} \cdot \mathbf{z}^{l-1}}^{T} + \mathbf{B}^l_0,
    \label{eqn:latent_matrix}
\end{equation}
where $[T]$ and $[\cdot]$ denote matrix transpose and multiplication, $\mathbf{Z}^l, \mathbf{B}_0 \in \mathbb{R}^{n \times c}$. Then every element in the latent matrix is projected to an $m$ dimensional embedding space, namely, 
\begin{equation}
    \mathbf{E}^{l}_{i,j} = \mathbf{Z}^{l}_{i,j} \mathbf{w}^l_1 + \mathbf{b}^l_1, i = 1, \cdots, n, j = 1, \cdots, c,
    \label{eqn:embedding}
\end{equation}
where $\mathbf{E}^{l}_{i,j}, \mathbf{w}^l_1, \mathbf{b}^l_1 \in \mathbb{R}^m$. The vectors $\mathbf{w}^l_1$ and $\mathbf{b}^l_1$ are element-wise unique and for the simplicity of notation, the subscript $_{i,j}$ is omitted. $\mathbf{w}^l_1$, $\mathbf{b}^l_1$, and $\mathbf{E}^{l}_{i,j}$ can be aggregated as 3D tensors, namely $\mathbf{W}^l_1, \mathbf{B}^l_1, \mathbf{E}^{l}  \in \mathbb{R}^{n \times c \times m}$.
After the operation in Eqn.~\ref{eqn:embedding}, the elements of $\mathbf{Z}^{l}$ are converted to embedded vectors in the embedding space. 
The final step is to obtain the output that can be explicitly used as the weights of the convolutional layer.
To achieve that, every embedded vector $\mathbf{E}^{l}_{i,j}$ is multiplied by an explicit matrix, that is,
\begin{equation}
    \mathbf{O}^l_{i,j} =  \mathbf{w}^l_2 \cdot \mathbf{E}^{l}_{i,j} + \mathbf{b}^l_2, i = 1, \cdots, n, j = 1, \cdots, c,
    \label{eqn:explicit}
\end{equation}
where $\mathbf{O}^l_{i,j}, \mathbf{b}^l_2 \in \mathbb{R}^{wh}$, $\mathbf{w}^l_2 \in \mathbb{R}^{wh \times m}$. Again, $\mathbf{w}^l_2$ and $\mathbf{b}^l_2$ are unique for every embedded vector and the subscript $_{i,j}$ is omitted.
$\mathbf{w}^l_2$, $\mathbf{b}^l_2$, and $\mathbf{O}^l_{i,j}$ can also be aggregated as high-dimensional tensors, \ie $\mathbf{W}^l_2 \in \mathbb{R}^{n \times c \times wh \times m}$ and $\mathbf{B}^l_2, \mathbf{O}^l \in \mathbb{R}^{n \times c \times wh}$. For the sake of simplicity, Eqn.~\ref{eqn:latent_matrix}, \ref{eqn:embedding} and \ref{eqn:explicit} can be abstracted as
\begin{equation}
    \mathbf{O}^l = h(\mathbf{z}^l, \mathbf{z}^{l-1}; \mathbf{W}^l, \mathbf{B}^l),
    \label{eqn:hypernetwork}
\end{equation}
where $h(\cdot)$ denotes the functionality of the hypernetwork.
The final output $\mathbf{O}^l$ is used as the weight parameter of the $l$-th layer. The output $\mathbf{O}^l$ is covariant with the input latent vector because pruning an element in the latent vector removes the corresponding slice of the output $\mathbf{O}^l$ (See Fig.~\ref{fig:hypernetwork_design}). 

When designing the hypernetwork, we tried to add batch normalization and non-linear layers after the linear operation in Eqn.~\ref{eqn:embedding} and Eqn.~\ref{eqn:explicit}. 
But it did not lead to clearly better results. 
Thus, we just kept to the simple design. This is also consistent with the previous designs~\cite{ha2017hypernetworks,liu2019metapruning}.

\subsection{Sparsity regularization and proximal gradient}
\label{subsec:sparsity_regularization}

The core of differentiability comes with not only the specifically designed hypernetwork but also the mechanism used to search the the potential candidate. To achieve that, we enforce sparsity constraints to the latent vectors. The loss function of the aforementioned $L$-layer CNN is denoted as
\begin{equation}
    \min_{\mathbf{W},\mathbf{B},\mathbf{z}}\mathcal{L}\Big(\mathbf{y}, f\big(\mathbf{x}; h(\mathbf{z}; \mathbf{W}, \mathbf{B})\big)\Big) +  \gamma \mathcal{D}(\mathbf{W}) + \gamma \mathcal{D}(\mathbf{B}) + \lambda \mathcal{R}(\mathbf{z}),
    \label{eqn:loss_function}
\end{equation}
where $\mathcal{L}(\cdot, \cdot)$, $\mathcal{D}(\cdot)$, and $\mathcal{R}(\cdot)$ are the loss function for a specific vision task, the weight decay term, and the sparsity regularization term, $\gamma$ and $\lambda$ are the regularization factors. For the simplicity of notation, the superscript $l$ is omitted. The sparsity regularization takes the form of $\ell_1$ norm, namely,
\begin{equation}
    \mathcal{R}(\mathbf{z}) = \sum_{l=1}^{L}\|\mathbf{z}^l\|_1.
    \label{eqn:sparsity_regularization}
\end{equation}

To solve the problem in Eqn.~\ref{eqn:loss_function}, the weights $\mathbf{W}$ and and biases $\mathbf{B}$ of the hypernetworks are updated with SGD. 
The gradients are back-propagated from the backbone network to the hypernetwork. 
Thus, neither the forward nor the backward pass challenges the information flow between them. 
The latent vectors are updated with proximal gradient algorithm, \ie,
\begin{equation}
    \mathbf{z}{[k+1]} = \mathbf{prox_{\lambda\mu\mathcal{R}}}\Big(\mathbf{z}{[k]} - \lambda\mu\nabla\mathcal{L}\big(\mathbf{z}{[k]} \big) \Big),
    \label{eqn:proximal_gradient}
\end{equation}
where $\mu$ is the step size of proximal gradient and is set as the learning rate of SGD updates.
As can be seen in the equation, the proximal gradient update contains a gradient descent step and a proximal operation step.
When the regularizer has the form of $\ell_1$ norm, the proximal operator has closed-form solution, \ie
\begin{equation}
    \mathbf{z}{[k+1]} = \mathrm{sgn}\Big(\mathbf{z}{[k+\Delta]}\Big)\Big[\big|\mathbf{z}{[k+\Delta]}\big| - \lambda\mu\Big]_+,
    \label{eqn:soft_thresholding}
\end{equation}
where $\mathbf{z}{[k+\Delta]} = \mathbf{z}{[k]} - \lambda\mu\nabla\mathcal{L}(\mathbf{z}{[k]})$ is the intermediate SGD update, the sign operator $\mathrm{sgn}(\cdot)$, the thresholding operator $[\cdot]_+$, and the absolute value operator $|\cdot|$ act element-wise on the vector. Eqn.~\ref{eqn:soft_thresholding} is the soft-thresholding function. 

The latent vectors first get SGD updates along with the other parameters $\mathbf{W}$ and $\mathbf{B}$. 
Then the proximal operator is applied. 
Due to the use of SGD updates and the fact that the proximal operator has closed-form solution, we recognize the whole solution as approximately differentiable (although the $\ell_1$ norm is not differentiable at 0), which guarantees the fast convergence of the algorithm compared with reinforcement learning and evolutionary algorithm. 
The speed-up of proximal gradient lies in that instead of searching the best candidate among the total population it forces the solution towards the best sparse one.

The automation of pruning follows the way the sparsity applied in Eqn.~\ref{eqn:sparsity_regularization} and the proximal gradient solution. 
First of all, all latent vectors are regularized together without distinguishment between them. 
During the optimization, information and gradients flows fluently between the backbone network and the hypernetwork. 
The proximal gradient algorithm forces the potential elements of the latent vectors to approach zero quicker than the others without any human effort and interference in this process. 
The optimization stops immediately when the target compression ratio is reached. 
In total, there are only two additional hyper-parameters in the algorithm, \ie the sparsity regularization factor and the mask threshold $\tau$ in Subsec.~\ref{subsec:network_pruning}. 
Thus, running the algorithm is just like turning on the button, which enable the application of the algorithm to all of the CNNs without much interference of domain experts' knowledge.

\subsection{Network pruning}
\label{subsec:network_pruning}

Different from the fully connected layers, the proposed design of hypernetwork can adapt the dimension of the output according to that of the latent vectors. 
After the searching stage, sparse versions of the latent vectors are derived as $\mathbf{\overline{z}}^{l-1}$ and $\mathbf{\overline{z}}^{l}$. 
For those vectors, some of their elements are zero or approaching zero. 
Thus, 1-0 masks can by derived by comparing the sparse latent vectors with a predefined small threshold $\tau$, \ie $\mathbf{m}^{l} = \mathcal{T}\left(\mathbf{\overline{z}}^{l}, \tau \right)$, where the function $\mathcal{T}(\cdot)$ element-wise compares the latent vector with the threshold and returns 1 if the element is not smaller than $\tau$ and 0 otherwise. 
Then latent vector $\mathbf{\overline{z}}^{l}$ is pruned according to the mask $\mathbf{m}^{l}$. 
See Subsec.~\ref{subsec:implementation_consideration} for more analysis.

\subsection{Latent vector sharing}
\label{subsec:latent_vector_sharing}

Due to the existence of skip connections in residual networks such as ResNet, MobileNetV2, SRResNet, and EDSR, the residual blocks are interconnected with each other in the way that their input and output dimensions are related. 
Therefore, the skip connections are notoriously tricky to deal with.
But back to the design of the proposed hypernetwork, a quite simple and straightforward solution to this problem is to let the hypernetworks of the correlated layers share the same latent vector.
Note that the weight and bias parameters of the hypernetworks are not shared.
Thus, sharing latent vectors does not force the the correlated layers to be identical.
By automatically pruning the single latent vector, all of the relevant layers are pruned together. 
Actually, we first tried to use different latent vectors for the correlated layers and applied group sparsity to them. 
But the experimental results showed that this is not a good choice because this strategy shot lower accuracy than the latent vector sharing strategy. (See details in the Supplementary). 

\subsection{Discussion on the convergence property}
\label{subsec:discussion}

Compared with reinforcement learning and evolutionary algorithm, proximal gradient may not be the optimal solution for some problems. 
But as found by previous works~\cite{liu2019rethinking,liu2019metapruning}, automatic network pruning serves as an implicit searching method for the channel configuration of a network. 
In addition, the network is searched and trained from scratch in this paper. 
The important factor is the number of remaining channels of the convolutional layers in the network.
Thus, it is relatively not important which filter is pruned as long as the number of pruned channels are the same. 
This reduces the number of possible candidates by orders of magnitude. In this case, proximal gradient works quite well. 

\subsection{Implementation consideration}
\label{subsec:implementation_consideration}

\textbf{Compact representation of the hypernetwork.} Thanks to the default tensor operations in deep learning toolboxes~\cite{paszke2017automatic}, the operations in the hypernetwork could be represented in a compact form. 
The embedding operation in Eqn.~\ref{eqn:embedding} can be written as the following high-dimensional tensor operation
\begin{equation}
    \mathbf{E}^{l} = \mathcal{U}^3\left( \mathbf{Z}^{l} \right) \circ \mathbf{W}^l_1 + \mathbf{B}^l_1, 
    \label{eqn:embedding_tensor}
\end{equation}
where $\mathcal{U}^3( \mathbf{Z}^{l}) \in \mathbb{R}^{n \times c \times 1}$, $[\circ]$ denotes the broadcastable element-wise tensor multiplication, $\mathcal{U}^3(\cdot)$ inserts a third dimension for $\mathbf{Z}^{l}$. 
The operation in Eqn.~\ref{eqn:explicit} can be easily rewritten as batched matrix multiplication,
\begin{equation}
    \mathbf{O}^l =  \mathbf{W}^l_2 * \mathbf{E}^{l} + \mathbf{B}^l_2, 
    \label{eqn:explicit_tensor}
\end{equation}
where $[*]$ denotes batched matrix multiplication.

\textbf{Pruning analysis.} Analyzing the three layers of the hypernetworks together with the masked latent vectors leads to a direct impression on how the backbone layers are automatically pruned. 
That is,
\begin{align}
     \mathbf{\overline{O}}^l & = \mathbf{W}^l_2 * \left[ \mathcal{U}^3\left( (\mathbf{m}^{l} \circ \mathbf{\overline{z}}^{l}) \cdot ({\mathbf{m}^{l-1} \circ \mathbf{\overline{z}}^{l-1}})^{T} \right) \circ \mathbf{W}^l_1 \right] \label{eqn:joint_analysis_one} \\
     & = \mathbf{W}^l_2 * \left[\mathcal{U}^3 (\mathbf{m}^{l} \cdot {\mathbf{m}^{l-1}}^T) \circ \mathcal{U}^3(\mathbf{\overline{z}}^{l} \cdot {\mathbf{\overline{z}}^{l-1}}^{T}) \circ \mathbf{W}^l_1 \right] \label{eqn:joint_analysis_two} \\
     & = \mathcal{U}^3 (\mathbf{m}^{l} \cdot {\mathbf{m}^{l-1}}^T) \circ \left[\mathbf{W}^l_2 * \left(\mathcal{U}^3(\mathbf{\overline{z}}^{l} \cdot {\mathbf{\overline{z}}^{l-1}}^{T}) \circ \mathbf{W}^l_1 \right) \right] \label{eqn:joint_analysis_three}
\end{align}
The equality follows the broadcastability of the the operations $[\circ]$ and $[*]$.
As shown in the above equations, applying the masks on the latent vectors has the same effect of applying them on the final output. 
Note that in the above analysis the bias terms $\mathbf{B}^l_0$, $\mathbf{B}^l_1$, and $\mathbf{B}^l_2$ are omitted since they have a really small influence on the output of the hypernetwork. 
In conclusion, the final output can be pruned according to the same criterion for the latent vectors.

\textbf{Initialization of the hypernetwork.} All biases are initialized as zero, the latent vector with standard normal distribution, and $\mathbf{W}^l_1$ with Xaiver uniform~\cite{glorot2010understanding}.
The weight of the explicit layer $\mathbf{W}^l_2$ is initialized with Hyperfan-in which guarantees stable backbone network weights and fast convergence~\cite{chang2020principled}.

\begin{table}[!t]
    \caption{Results on image classification networks. The FLOPs ratio and parameter ratio of the pruned networks are reported. DHP outperforms the compared methods under comparable model complexity. On Tiny-ImageNet, DHP-24-2 shoots lower error rates than the original model}
    % \vspace{-0.1cm}
    \label{tbl:results_classification}
    \scriptsize
    \begin{center}
        \begin{tabular}{c|c|c|c|c}
            \toprule
            \multirow{2}{*}{\shortstack{Network \\ Top-1 Error (\%)} }  & Compression & \multirow{2}{*}{Top-1 Error (\%)} & \multirow{2}{*}{FLOPs Ratio (\%)} & \multirow{2}{*}{Parameter  Ratio (\%)} \\ 
             & Method &&& \\ \midrule
            \multicolumn{5}{c}{CIFAR10} \\ \midrule
            \multirow{3}{*}{\shortstack{ResNet-20 \\ 7.46}} 
            & \cite{ye2018rethinking}               & 9.10 & 52.60 & 62.78 \\
            & \textbf{DHP-50 (Ours)}                & 8.46 & 51.80 & 56.13 \\ 
            & FPGM~\cite{he2019filter}              & 9.38 & 46.00 & -- \\ \midrule
            
            \multirow{12}{*}{\shortstack{ResNet-56 \\ 7.05}} 
            & Variational~\cite{zhao2019variational}& 7.74 & 79.70 & 79.51 \\
            & Pruned-B~\cite{li2017pruning}         & 6.94 & 72.40 & 86.30 \\
            & GAL-0.6~\cite{lin2019towards}         & 6.62 & 63.40 & 88.20 \\
            & NISP~\cite{yu2018nisp}                & 6.99 & 56.39 & 57.40 \\
            & \textbf{DHP-50 (Ours)}                & 6.42 & 50.96 & 58.42 \\
            & CaP~\cite{Minnehan_2019_CVPR_CaP}     & 6.78 & 50.20 & -- \\
            & ENC~\cite{Kim_2019_CVPR_ENC}          & 7.00 & 50.00 & -- \\
            & AMC~\cite{He_2018_ECCV_AMC}           & 8.10 & 50.00 & -- \\
            & KSE~\cite{Li_2019_CVPR_KSE}           & 6.77 & 48.00 & 45.27 \\
            & FPGM~\cite{he2019filter}              & 6.74 & 47.70 & -- \\  
            & GAL-0.8~\cite{lin2019towards}         & 8.42 & 39.80 & 34.10 \\
            & \textbf{DHP-38 (Ours)}                & 7.06 & 39.07 & 41.10  \\ \midrule
            
            \multirow{4}{*}{\shortstack{ResNet-110 \\ 5.31}}
            & \textbf{DHP-62 (Ours)}                & 5.37 & 63.66 & 63.2   \\
            & Variational~\cite{zhao2019variational}& 7.04 & 63.56 & 58.73  \\
            & Pruned-B~\cite{li2017pruning}         & 6.70 & 61.40 & 67.60  \\
            & GAL-0.5~\cite{lin2019towards}         & 7.26 & 51.50 & 55.20  \\ 
            & \textbf{DHP-20 (Ours)}                & 6.61 & 21.63 & 22.40  \\ \midrule
            
            \multirow{4}{*}{\shortstack{ResNet-164 \\ 4.97}}
            & Hinge~\cite{li2020group}              & 5.40 & 53.61 & 70.34  \\
            & SSS~\cite{huang2018data}              & 5.78 & 53.53 & 84.75  \\
            & \textbf{DHP-50 (Ours)}                & 5.22 & 51.67 & 50.97  \\ 
            & Variational~\cite{zhao2019variational}& 6.84 & 50.92 & 43.30  \\ 
            & \textbf{DHP-20 (Ours)}                & 6.30 & 21.78 & 20.46  \\    \midrule
            
            \multirow{3}{*}{\shortstack{DenseNet-12-40 \\ 5.26}} 
            & Variational~\cite{zhao2019variational}& 6.84 & 55.22 & 40.33  \\ 
            & \textbf{DHP-38 (Ours)}                & 6.06 & 39.80 & 63.76  \\             
            & \textbf{DHP-28 (Ours)}                & 6.51 & 29.52 & 26.01  \\
            & GAL-0.1~\cite{lin2019towards}         & 6.77 & 28.60 & 25.00  \\  \midrule
            
            \multicolumn{5}{c}{Tiny-ImageNet} \\ \midrule
            \multirow{4}{*}{\shortstack{MobileNetV1 \\ 52.71}} 
            & \textbf{DHP-24-2 (Ours)}              & 50.75  & 101.08 & 43.58  \\
            & MobileNetV1-0.75                      & 54.22  & 57.42  & 57.64  \\ 
            & MetaPruning~\cite{liu2019metapruning} & 54.48  & 56.77  & 88.14  \\ 
            & \textbf{DHP-50 (Ours)}                & 51.63  & 51.91  & 36.95  \\ \midrule
            
            \multirow{4}{*}{\shortstack{MobileNetV2 \\ 44.75}} 
            & \textbf{DHP-24-2 (Ours)}              & 43.82  & 99.09  & 72.72  \\
            & \textbf{DHP-10 (Ours)}                & 52.43  & 11.92  & 6.50   \\ 
            & MetaPruning~\cite{liu2019metapruning} & 56.72  & 11.00  & 90.27  \\ 
            & MobileNetV2-0.3                       & 53.99  & 10.09  & 11.64  \\ \bottomrule
        \end{tabular}
    \end{center}
    % \vspace{-0.4cm}
\end{table}

\section{Experimental Results}
\label{sec:experimental_results}

% \textbf{Experiment setup.} 
% 
To validate the proposed method, extensive experiments were conducted on various CNN architectures including ResNet~\cite{he2016deep}, DenseNet~\cite{huang2017densely} for CIFAR10~\cite{krizhevsky2009learning} image classification, MobileNetV1~\cite{howard2017mobilenets}, MobileNetV2~\cite{sandler2018mobilenetv2} for Tiny-ImageNet~\cite{deng2009imagenet} image classification, SRResNet~\cite{ledig2016photo}, EDSR~\cite{lim2017enhanced} for single image super-resolution, and DnCNN~\cite{zhang2017beyond}, UNet~\cite{ronneberger2015unet} for gray image denoising. 
The proposed DHP algorithm starts from a randomly initialized network with the initialization method detailed in Subsec.~\ref{subsec:implementation_consideration}. 
After pruning, the training of the pruned network continues with the same training protocol used for the original network.
All of the experiments are conducted on NVIDIA TITAN Xp GPUs. 

\textbf{Hyperparameters.} 
A target FLOPs compression ratio is set for the pruning algorithm. 
When the difference between the target compression ratio and the actual compression ratio falls below 2\%, the automatic pruning procedure stops.
The parameter space of hypernetwork increases in proportion to the dimension $m$ of the embedding space. Thus, $m$ should not be too large. In this paper, $m$ is set to 8. 
The step size $\mu$ of the proximal operator is set as the learning rate of SGD updates. 
The sparsity regularization factor $\lambda$ is set by empirical studies. 
The value is chosen such that the searching epochs constitute arounds 5\% -- 10\% of the whole training epochs. 
This guarantees acceptable convergence during searching while not introducing too much additional computation.
Please refer to the supplementary for the detailed training and testing protocol.  

\subsection{Image classification}
\label{subsec:image_classification}

The compression results on image classification networks are shown in Table~\ref{tbl:results_classification}. 
`DHP-**' denotes the proposed method with the target FLOPs ratio during pruning stage. 
As in Fig.~\ref{fig:mobilenet_dhp_comparison}, the operating point DHP-24-2 is derived by compressing the widened mobile networks with $\alpha = 2$ and the target FLOPs ratio 24\%.
For ResNet-56, the proposed method is compared with 9 different network compression methods and achieves the best performance, \ie 6.42\% Top-1 error rate on the most intensively investigated 50\% compression level. 
The compression of DenseNet-12-40 is reasonable compared with the other methods. 
The accuracy of the operating points DHP-62 of ResNet-110 and DHP-50 of ResNet-164 is quite close to that of the baseline. 
More results on ResNet-110 and ResNet-164 are shown in the Supplementary. 
A comparison between $\ell_1$ and $\ell_2$ regularization in Eqn.~\ref{eqn:sparsity_regularization} is done. 
The results in the Supplementary shows that $\ell_1$ regularization is better than $\ell_2$ regularization.

On Tiny-ImageNet, DHP achieves lower Top-1 error rates than MetaPruning~\cite{liu2019metapruning} under the same FLOPs constraint.
DHP results in models that are more accurate than the uniformly scaled networks. 
The error rate of DHP-10 is 1.56\% lower than that of MobileNetV2-0.3 with slightly fewer FLOPs and 5.14\% fewer parameters.
On MobileNetV1, the accuracy gain of DHP-50 over MobileNetV1-0.75 goes to 2.59\% with over 5\% fewer FLOPs and 10\% fewer parameters. 
Based on this, we hypothesized that it is possible to derive a model which is more accurate than the original version by pruning the widened mobile networks. 
And this is confirmed by comparing the accuracy of the operating points DHP-24-2 with the baseline accuracy.

\subsection{Super-resolution}
\label{subsec:super_resolution}

\begin{figure*}[!t]\footnotesize%\small
	%\centering
    % \hspace{-0.23cm}
    \resizebox{\linewidth}{!}
    {
    \begin{tabular}{c@{\extracolsep{0.0em}}c@{\extracolsep{0.0em}}c@{\extracolsep{0.0em}}c@{\extracolsep{0.0em}}c@{\extracolsep{0.0em}}c}
		\includegraphics[width=0.255\textwidth]{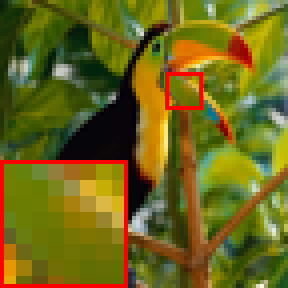}~
        &\includegraphics[width=0.255\textwidth]{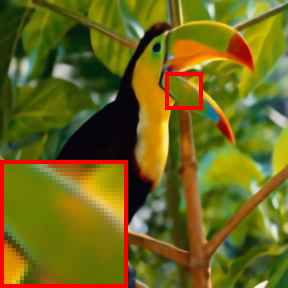}~
		&\includegraphics[width=0.255\textwidth]{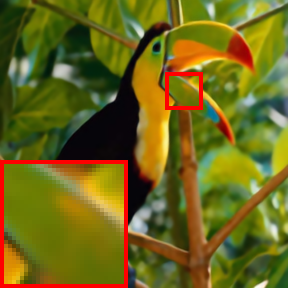}~
		&\includegraphics[width=0.255\textwidth]{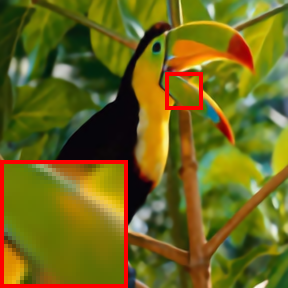}~
		&\includegraphics[width=0.255\textwidth]{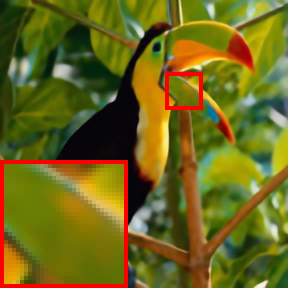}~
		&\includegraphics[width=0.255\textwidth]{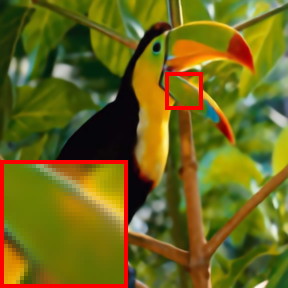}\\
        PSNR/FLOPs/Runtime  & 32.85/28.59/14.10     & 32.50/28.59/19.75     & 32.65/19.82/14.71     & 32.24/19.28/25.49     & 32.64/17.61/5.40  \\
		(a) LR              & (b) EDSR              & (d)  Cluster          & (c) Basis             & (f) Factor            & (e) DHP           \\
	\end{tabular}
	}
% 	\vspace{-0.1cm}
	\caption{Single image super-resolution visual results. PSNR and FLOPs measured on the image. Runtime averaged on Set5}
% 	\vspace{-0.2cm}
	\label{fig:super_resolution_visual}
\end{figure*}

\begin{table}[t]
    % \small
    % \vspace{-0.1cm}
    \caption{Results on image super-resolution networks. The upscaling factor is $\times 4$. Runtime is averaged for Urban100. Maximum GPU memory consumption is reported for Urban100. FLOPs is reported for a $128 \times 128$ image patch. DHP achieves significant reduction of runtime}
    % \vspace{-0.1cm}
    \label{tbl:results_super_resolution}
    \scriptsize
    \begin{center}
    \resizebox{\linewidth}{!}
    {
        \begin{tabular}{c|c|c|c|c|c|c|c|r|r|c}
            \toprule
            \multirow{3}{*}{Network} & \multirow{3}{*}{Method} & \multicolumn{5}{|c|}{\multirow{2}{*}{PSNR $[dB]$}} & \multirow{3}{*}{\shortstack[c]{FLOPs \\ $[G]$}} & \multirow{3}{*}{\shortstack[c]{Params \\ $[M]$}} & \multirow{3}{*}{\shortstack[c]{Run- \\ time \\ $[ms]$}} & \multirow{3}{*}{\shortstack[c]{GPU \\ Mem \\ $[GB]$}} \\ 
            & & \multicolumn{5}{|c|}{} & & & &\\ \cline{3-7} 
            &  & Set5 & Set14 & B100 & Urban100 & DIV2K &&&& \\ \midrule
            \multirow{8}{*}{SRResNet} & Baseline            & 32.03  & 28.50  & 27.52  & 25.88  & 28.85  & 32.83  & 1.54  & 34.73  & 0.6773  \\
            \multirow{8}{*}{\cite{ledig2016photo}}& 
            Clustering~\cite{son2018clustering}             & 31.93  & 28.44  & 27.47  & 25.71  & 28.75  & 32.83  & 0.34  & 31.07  & 0.8123  \\            
            & Factor-SIC3~\cite{wang2017factorized}         & 31.86  & 28.38  & 27.40  & 25.58  & 28.65  & 20.83  & 0.81  & 102.51  & 1.4957  \\
            & \textbf{DHP-60  (Ours)}                       & 31.97  & 28.47  & 27.48  & 25.76  & 28.79  & 20.29  & 0.95  & 27.91    & 0.5923  \\    
            & Basis-32-32~\cite{li2019learning}             & 31.90  & 28.42  & 27.44  & 25.65  & 28.69  & 19.77  & 0.74  & 45.73   & 0.9331  \\
            & Factor-SIC2~\cite{wang2017factorized}         & 31.68  & 28.32  & 27.37  & 25.47  & 28.58  & 18.38  & 0.66  & 74.66   & 1.1201  \\
            & Basis-64-14~\cite{li2019learning}             & 31.84  & 28.38  & 27.39  & 25.54  & 28.63  & 17.49  & 0.60  & 36.75   & 0.6741  \\ 
            & \textbf{DHP-40  (Ours)}                       & 31.90  & 28.45  & 27.47  & 25.72  & 28.75  & 13.71  & 0.64  & 22.71  & 0.4907  \\ 
            & \textbf{DHP-20  (Ours)}                       & 31.77  & 28.34  & 27.40  & 25.55  & 28.60  & 7.77   & 0.36  & 14.74  & 0.3795  \\             
            \midrule
            \multirow{8}{*}{EDSR}   & Baseline              & 32.10  & 28.55  & 27.55  & 26.02  & 28.93 & 90.37  &  3.70 & 49.73  & 1.3276 \\ 
            \multirow{8}{*}{\cite{lim2017enhanced}}
            & Clustering~\cite{son2018clustering}           & 31.93  & 28.47  & 27.48  & 25.77  & 28.80  & 90.37 & 0.82  & 50.51  & 1.2838 \\ 
            & Factor-SIC3~\cite{wang2017factorized}         & 31.96  & 28.47  & 27.49  & 25.81  & 28.81  & 65.49 & 2.19 & 125.10  & 1.5007 \\
            & Basis-128-40~\cite{li2019learning}            & 32.03  & 28.45  & 27.50  & 25.81  & 28.82  & 62.65 & 2.00 & 48.19  & 1.3219 \\
            & Factor-SIC2~\cite{wang2017factorized}         & 31.82  & 28.40  & 27.43  & 25.63  & 28.70  & 60.90 & 1.90 & 94.94  & 1.3209 \\
            & Basis-128-27~\cite{li2019learning}            & 31.95  & 28.42  & 27.46  & 25.76  & 28.76  & 58.28 & 1.74 & 45.84  & 1.3209\\
            & \textbf{DHP-60 (Ours)}                        & 31.99  & 28.52  & 27.53  & 25.92  & 28.88  & 55.67 & 2.28 & 45.11  & 0.6950\\  
            & \textbf{DHP-40 (Ours)}                        & 32.01  & 28.49  & 27.52  & 25.86  & 28.85  & 37.77 & 1.53  & 33.50  & 0.9650\\ 
            & \textbf{DHP-20 (Ours)}                        & 31.94  & 28.42  & 27.47  & 25.69  & 28.77  & 19.40 & 0.79   & 22.63  & 1.1588\\
            \bottomrule
        \end{tabular}
    }
    \end{center}
    % \vspace{-0.4cm}
\end{table}
The results on image super-resolution networks are shown in Table~\ref{tbl:results_super_resolution}. 
DHP is compared with factorized convolution (Factor)~\cite{wang2017factorized}, learning filter basis method (Basis)~\cite{li2019learning}, and K-means clustering method (Clustering)~\cite{son2018clustering}. 
`SIC*' denotes the number of SIC layers in Factor~\cite{wang2017factorized}. 
The practical FLOPs instead of the theoretical FLOPs is reported for Clustering~\cite{son2018clustering}. 
To fairly compare the methods and measure the practical compression effectiveness, five metrics are involved including Peak Signal-to-Noise Ratio (PSNR), FLOPs, number of parameters, runtime and GPU memory consumption. 
Several conclusion can be drawn. \textbf{I.} Previous methods mainly focus on the reduction of FLOPs and number of parameter without paying special attention to the actual acceleration. Although Clustering can reduce substantial parameters while maintaining quite good PSNR accuracy, the actual computing resource requirement (GPU memory and runtime) is remained. \textbf{II.} Convolution factorization and decomposition methods result in additional CUDA kernel calls, which is not efficient for the actual acceleration. \textbf{III.} For the proposed method, the two model complexity metrics, \ie FLOPs and parameters change consistently across different operating points, which leads to consistent reduction of computation resources. \textbf{IV.} DHP results in both inference-efficient (DHP-20) and accuracy-preserving (DHP-60) models. The visual results are shown in Fig.~\ref{fig:super_resolution_visual}. As can been seen, the visual quality of the images of DHP is almost indistinguishable from that of the baseline.

\subsection{Denoising}
\label{subsec:denoising}
\begin{figure*}[!t]\footnotesize%\small
	%\centering
    % \hspace{-0.23cm}
    \resizebox{\linewidth}{!}
    {
    \begin{tabular}{c@{\extracolsep{0.0em}}c@{\extracolsep{0.0em}}c@{\extracolsep{0.0em}}c@{\extracolsep{0.0em}}c@{\extracolsep{0.0em}}c}
		\includegraphics[width=0.255\textwidth]{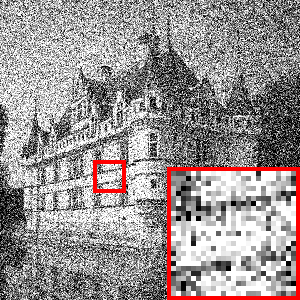}~
		&\includegraphics[width=0.255\textwidth]{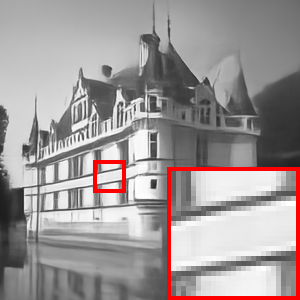}~
		&\includegraphics[width=0.255\textwidth]{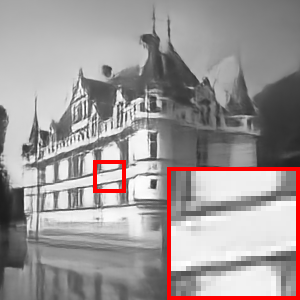}~
		&\includegraphics[width=0.255\textwidth]{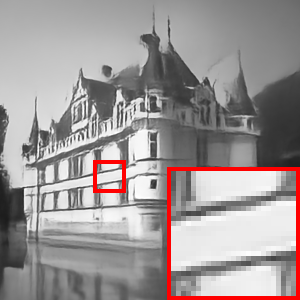}~
        &\includegraphics[width=0.255\textwidth]{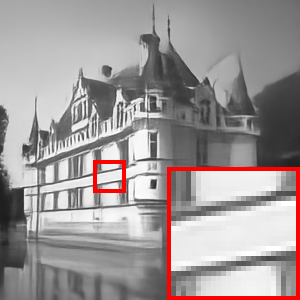}~
        &\includegraphics[width=0.255\textwidth]{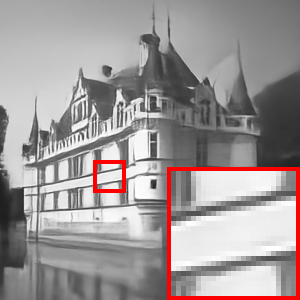}		\\
        PSNR/FLOPs/Runtime  & 25.60/1.08/7.27   & 25.30/1.08/9.66   &25.37/0.49/40.36   & 25.51/0.47/9.00   & 25.57/0.45/6.09   \\
		(a) Noisy           & (b) UNet          & (f) Cluster       & (d) Factor        & (e) Group         & (c) DHP        \\
	\end{tabular}
	}
% 	\vspace{-0.1cm}
	\caption{Image denoising visual results. PSNR and FLOPs measured on the image. Runtime averaged on B100}
% 	\vspace{-0.2cm}
	\label{fig:denoising_visual}
\end{figure*}

\begin{table}[t]
    % \small
    \caption{Results on image denoising networks. The noise level is 70. 
    Runtime and maximum GPU memory are reported for BSD68. FLOPs is reported for a $128 \times 128$ image. DHP achieves significant reduction of runtime} 
    % \vspace{-0.1cm}
    \scriptsize
    \begin{center}
        \begin{tabular}{c|c|c|c|c|c|r|c}
            \toprule
            \multirow{2}{*}{Network} & \multirow{2}{*}{Method} & \multicolumn{2}{|c|}{{PSNR $[dB]$}} & \multirow{2}{*}{\shortstack[c]{FLOPs \\ $[G]$}} & \multirow{2}{*}{\shortstack[c]{Params \\ $[M]$}} & \multirow{2}{*}{\shortstack[c]{Runtime \\ $[ms]$}} & \multirow{2}{*}{\shortstack[c]{GPU Mem \\ $[GB]$}} \\ \cline{3-4} 
            % & & \multicolumn{2}{|c|}{} & & & &\\ \cline{3-4} 
            &  & BSD68 & DIV2K &&&& \\ \midrule
            \multirow{8}{*}{DnCNN~\cite{zhang2017beyond}}   & Baseline  & 24.93  & 26.73  & 9.13  & 0.56  & 23.38 & 0.1534  \\ 
            & Clustering~\cite{son2018clustering}           & 24.90  & 26.67  & 9.13  & 0.12  & 21.97  & 0.2973  \\  
            & \textbf{DHP-60 (Ours)}                        & 24.91  & 26.69  & 5.65  & 0.34  & 18.90  & 0.1443 \\ 
            & \textbf{DHP-40 (Ours)}                        & 24.89  & 26.65  & 3.83  & 0.23  & 14.62  & 0.1194  \\ 
            & Factor-SIC3~\cite{wang2017factorized}         & 24.97  & 26.83  & 3.54  & 0.22  & 125.46  & 0.5910  \\ 
            & Group~\cite{peng2018extreme}                  & 24.88  & 26.64  & 3.34  & 0.20  & 25.69  & 0.1807  \\ 
            & Factor-SIC2~\cite{wang2017factorized}         & 24.93  & 26.76  & 2.38  & 0.15  & 84.17 & 0.4149  \\             
            & \textbf{DHP-20 (Ours)}                        & 24.84  & 26.58  & 2.01  & 0.12  & 10.72  & 0.0869  \\ 
            \midrule
            \multirow{8}{*}{UNet~\cite{ronneberger2015unet}}& Baseline  & 25.17  & 27.17  & 3.41  & 7.76  & 8.73 & 0.1684  \\ 
            & Clustering~\cite{son2018clustering}           & 25.01  & 26.90  & 3.41  & 1.72  & 10.01  & 0.6704  \\ 
            & \textbf{DHP-60 (Ours)}                        & 25.14  & 27.11  & 2.11  & 4.76  & 6.86  & 0.4992  \\ 
            & Factor-SIC3~\cite{wang2017factorized}         & 25.04  & 26.94  & 1.56  & 3.42  & 39.84  & 0.1889  \\ 
            & Group~\cite{peng2018extreme}                  & 25.13  & 27.08  & 1.49  & 2.06  & 11.20  & 0.1481  \\ 
            & \textbf{DHP-40 (Ours)}                        & 25.12  & 27.08  & 1.43  & 3.24  & 4.50  & 0.4992  \\ 
            & Factor-SIC2~\cite{wang2017factorized}         & 25.01  & 26.90  & 1.22  & 2.51  & 30.16  & 0.1855  \\             
            & \textbf{DHP-20 (Ours)}                        & 25.04  & 26.97  & 0.75  & 1.61  & 3.93  & 0.4992  \\ 
            \bottomrule
        \end{tabular}
    \end{center}
    \label{tbl:results_denoising}
    % \vspace{-0.4cm}
\end{table}
The results for image denoising networks are shown in Table~\ref{tbl:results_denoising}. 
The same metrics as super-resolution are reported for denoising. 
An additional method, \ie filter group approximation (Group)~\cite{peng2018extreme} is included. 
In addition to the same conclusion in Subsec.~\ref{subsec:super_resolution}, another two conclusions are drawn here.
\textbf{I.} Group~\cite{peng2018extreme} fails to reduce the actual computation resources although with quite good accuracy and satisfactory reduction of FLOPs and number of parameters. 
This might due to the additional $1\times1$ convolution and possibly the inefficient implementation of group convolution in current deep learning toolboxes. 
\textbf{II.} For DnCNN, one interesting phenomenon is that Factor~\cite{wang2017factorized} is more accurate than the baseline but has larger appetite for other resources. 
This is due to two facts. Firstly, Factor~\cite{wang2017factorized} has skip connections within the SIC layer. 
The higher accuracy of Factor~\cite{wang2017factorized} just validates the effectiveness of skip connections.
Secondly, the SIC layer of Factor~\cite{wang2017factorized} introduces more convolutional layers. 
So Factor-SIC3 has five times more convolutioinal layers than the baseline, which definitely slows down the execution. % 
The visual results are shown in Fig.~\ref{fig:denoising_visual}. 
~
\section{Conclusion and Future Work}
\label{sec:conlusion}

In this paper, we proposed a differentiable automatic meta pruning method via hypernetwork for network compression. The differentiability comes with the specially designed hypernetwork and the proximal gradient used to search the potential candidate network configurations. The automation of pruning lies in the uniformly applied $\ell_1$ sparsity on the latent vectors and the proximal gradient that solves the problem. By pruning mobile network with width multiplier $\alpha = 2$, we obtained models with higher accuracy but lower computation complexity than that with $\alpha = 1$. We hypothesize this is due to the per-layer distinguishing configuration resulting from the automatic pruning. Future work might be investigating whether this phenomenon reoccurs for the other networks.

\section*{Acknowledgements}
This work was partly supported by the ETH Z\"urich Fund (OK), a Huawei Technologies Oy (Finland) project, an Amazon AWS grant, and an Nvidia grant.

\bibliographystyle{splncs04}
\bibliography{ms}

\begin{thebibliography}{10}
\providecommand{\url}[1]{\texttt{#1}}
\providecommand{\urlprefix}{URL }
\providecommand{\doi}[1]{https://doi.org/#1}

\bibitem{mitreview2020}
{MIT} technology review: 10 breakthrough technologies 2020.
  \url{https://www.technologyreview.com/lists/technologies/2020/#tiny-ai},
  accessed: 2020-03-01

\bibitem{Agustsson_2017_CVPR_Workshops}
Agustsson, E., Timofte, R.: {NTIRE} 2017 challenge on single image
  super-resolution: Dataset and study. In: Proc. CVPRW (July 2017)

\bibitem{alvarez2016learning}
Alvarez, J.M., Salzmann, M.: Learning the number of neurons in deep networks.
  In: Proce. NeurIPS. pp. 2270--2278 (2016)

\bibitem{alvarez2017compression}
Alvarez, J.M., Salzmann, M.: Compression-aware training of deep networks. In:
  Proc. NeurIPS. pp. 856--867 (2017)

\bibitem{bevilacqua2012low}
Bevilacqua, M., Roumy, A., Guillemot, C., Alberi-Morel, M.L.: Low-complexity
  single-image super-resolution based on nonnegative neighbor embedding. In:
  Proc. BMVC (2012)

\bibitem{brock2018smash}
Brock, A., Lim, T., Ritchie, J., Weston, N.: {SMASH}: One-shot model
  architecture search through hypernetworks. In: Proc. ICLR (2018)

\bibitem{chang2020principled}
Chang, O., Flokas, L., Lipson, H.: Principled weight initialization for
  hypernetworks. In: Proc. ICLR (2020)

\bibitem{chen2018constraint}
Chen, C., Tung, F., Vedula, N., Mori, G.: Constraint-aware deep neural network
  compression. In: Proc. ECCV. pp. 400--415 (2018)

\bibitem{chin2020towards}
Chin, T.W., Ding, R., Zhang, C., Marculescu, D.: Towards efficient model
  compression via learned global ranking. In: Proc. CVPR. pp. 1518--1528 (2020)

\bibitem{deng2009imagenet}
Deng, J., Dong, W., Socher, R., Li, L.J., Li, K., Fei-Fei, L.: Image{N}et: A
  large-scale hierarchical image database. In: Proc. CVPR. pp. 248--255. IEEE
  (2009)

\bibitem{dong2017learning}
Dong, X., Chen, S., Pan, S.: Learning to prune deep neural networks via
  layer-wise optimal brain surgeon. In: Proc. NIPS. pp. 4857--4867 (2017)

\bibitem{glorot2010understanding}
Glorot, X., Bengio, Y.: Understanding the difficulty of training deep
  feedforward neural networks. In: Proc. AISTATS. pp. 249--256 (2010)

\bibitem{ha2017hypernetworks}
Ha, D., Dai, A., Le, Q.V.: Hyper{N}etworks. In: Proc. ICLR (2017)

\bibitem{han2015deep}
Han, S., Mao, H., Dally, W.J.: Deep compression: Compressing deep neural
  networks with pruning, trained quantization and {H}uffman coding. In: Proc.
  ICLR (2015)

\bibitem{hassibi1993second}
Hassibi, B., Stork, D.G.: Second order derivatives for network pruning: Optimal
  brain surgeon. In: Proc. NeurIPS. pp. 164--171 (1993)

\bibitem{hayashi2019einconv}
Hayashi, K., Yamaguchi, T., Sugawara, Y., Maeda, S.i.: Einconv: Exploring
  unexplored tensor decompositions for convolutional neural networks. In: Proc.
  NeurIPS. pp. 5553--5563 (2019)

\bibitem{he2016deep}
He, K., Zhang, X., Ren, S., Sun, J.: Deep residual learning for image
  recognition. In: Proc. CVPR. pp. 770--778 (2016)

\bibitem{he2019filter}
He, Y., Liu, P., Wang, Z., Hu, Z., Yang, Y.: Filter pruning via geometric
  median for deep convolutional neural networks acceleration. In: Proc. CVPR.
  pp. 4340--4349 (2019)

\bibitem{He_2018_ECCV_AMC}
He, Y., Lin, J., Liu, Z., Wang, H., Li, L.J., Han, S.: {AMC}: Auto{ML} for
  model compression and acceleration on mobile devices. In: Proc. ECCV. pp.
  784--800 (2018)

\bibitem{he2017channel}
He, Y., Zhang, X., Sun, J.: Channel pruning for accelerating very deep neural
  networks. In: Proc. ICCV. pp. 1389--1397 (2017)

\bibitem{howard2017mobilenets}
Howard, A.G., Zhu, M., Chen, B., Kalenichenko, D., Wang, W., Weyand, T.,
  Andreetto, M., Adam, H.: Mobile{N}ets: Efficient convolutional neural
  networks for mobile vision applications. arXiv preprint arXiv:1704.04861
  (2017)

\bibitem{huang2017densely}
Huang, G., Liu, Z., van~der Maaten, L., Weinberger, K.Q.: Densely connected
  convolutional networks. In: Proc. CVPR. pp. 2261--2269 (2017)

\bibitem{Huang-CVPR-2015}
Huang, J.B., Singh, A., Ahuja, N.: Single image super-resolution from
  transformed self-exemplars. In: Proc. CVPR. pp. 5197--5206 (2015)

\bibitem{huang2018data}
Huang, Z., Wang, N.: Data-driven sparse structure selection for deep neural
  networks. In: Proc. ECCV. pp. 304--320 (2018)

\bibitem{Kim_2019_CVPR_ENC}
Kim, H., Umar Karim~Khan, M., Kyung, C.M.: Efficient neural network
  compression. In: Proc. CVPR (June 2019)

\bibitem{krizhevsky2009learning}
Krizhevsky, A., Hinton, G.: Learning multiple layers of features from tiny
  images. Tech. rep., Citeseer (2009)

\bibitem{krueger2017bayesian}
Krueger, D., Huang, C.W., Islam, R., Turner, R., Lacoste, A., Courville, A.:
  Bayesian hypernetworks. arXiv preprint arXiv:1710.04759  (2017)

\bibitem{lecun1990optimal}
LeCun, Y., Denker, J.S., Solla, S.A.: Optimal brain damage. In: Proc. NeurIPS.
  pp. 598--605 (1990)

\bibitem{ledig2016photo}
Ledig, C., Theis, L., Husz{\'a}r, F., Caballero, J., Cunningham, A., Acosta,
  A., Aitken, A., Tejani, A., Totz, J., Wang, Z., et~al.: Photo-realistic
  single image super-resolution using a generative adversarial network. In:
  Proc. CVPR. pp. 105--114 (2017)

\bibitem{li2017pruning}
Li, H., Kadav, A., Durdanovic, I., Samet, H., Graf, H.P.: Pruning filters for
  efficient convnets. In: Proc. ICLR (2017)

\bibitem{li2019oicsr}
Li, J., Qi, Q., Wang, J., Ge, C., Li, Y., Yue, Z., Sun, H.: {OICSR}:
  Out-in-channel sparsity regularization for compact deep neural networks. In:
  Proc. CVPR. pp. 7046--7055 (2019)

\bibitem{li2020group}
Li, Y., Gu, S., Mayer, C., Van~Gool, L., Timofte, R.: Group sparsity: The hinge
  between filter pruning and decomposition for network compression. In: Proc.
  CVPR (2020)

\bibitem{li2019learning}
Li, Y., Gu, S., Van~Gool, L., Timofte, R.: Learning filter basis for
  convolutional neural network compression. In: Proc. ICCV. pp. 5623--5632
  (2019)

\bibitem{Li_2019_CVPR_KSE}
Li, Y., Lin, S., Zhang, B., Liu, J., Doermann, D., Wu, Y., Huang, F., Ji, R.:
  Exploiting kernel sparsity and entropy for interpretable {CNN} compression.
  In: Proc. CVPR (2019)

\bibitem{lim2017enhanced}
Lim, B., Son, S., Kim, H., Nah, S., Lee, K.M.: Enhanced deep residual networks
  for single image super-resolution. In: Proc. CVPRW. pp. 1132--1140 (2017)

\bibitem{lin2019towards}
Lin, S., Ji, R., Yan, C., Zhang, B., Cao, L., Ye, Q., Huang, F., Doermann, D.:
  Towards optimal structured cnn pruning via generative adversarial learning.
  In: Proc. CVPR. pp. 2790--2799 (2019)

\bibitem{liu2015sparse}
Liu, B., Wang, M., Foroosh, H., Tappen, M., Pensky, M.: Sparse convolutional
  neural networks. In: Proc. CVPR. pp. 806--814 (2015)

\bibitem{liu2018nas}
Liu, C., Zoph, B., Neumann, M., Shlens, J., Hua, W., Li, L.J., Fei-Fei, L.,
  Yuille, A., Huang, J., Murphy, K.: Progressive neural architecture search.
  In: Proc. ECCV (September 2018)

\bibitem{liu2019darts}
Liu, H., Simonyan, K., Yang, Y.: {DARTS}: Differentiable architecture search.
  In: Proc. ICLR (2019)

\bibitem{liu2019metapruning}
Liu, Z., Mu, H., Zhang, X., Guo, Z., Yang, X., Cheng, T.K.T., Sun, J.:
  Meta{P}runing: Meta learning for automatic neural network channel pruning.
  In: Proc. ICCV (2019)

\bibitem{liu2019rethinking}
Liu, Z., Sun, M., Zhou, T., Huang, G., Darrell, T.: Rethinking the value of
  network pruning. In: Proc. ICLR (2019)

\bibitem{MartinFTM01}
Martin, D., Fowlkes, C., Tal, D., Malik, J.: A database of human segmented
  natural images and its application to evaluating segmentation algorithms and
  measuring ecological statistics. In: Proc. ICCV. vol.~2, pp. 416--423 (July
  2001)

\bibitem{Minnehan_2019_CVPR_CaP}
Minnehan, B., Savakis, A.: Cascaded projection: End-to-end network compression
  and acceleration. In: Proc. CVPR (June 2019)

\bibitem{molchanov2019importance}
Molchanov, P., Mallya, A., Tyree, S., Frosio, I., Kautz, J.: Importance
  estimation for neural network pruning. In: Proc. CVPR. pp. 11264--11272
  (2019)

\bibitem{pan2018hyperst}
Pan, Z., Liang, Y., Zhang, J., Yi, X., Yu, Y., Zheng, Y.: Hyperst-net:
  Hypernetworks for spatio-temporal forecasting. arXiv preprint
  arXiv:1809.10889  (2018)

\bibitem{paszke2017automatic}
Paszke, A., Gross, S., Chintala, S., Chanan, G., Yang, E., DeVito, Z., Lin, Z.,
  Desmaison, A., Antiga, L., Lerer, A.: Automatic differentiation in
  {P}y{T}orch  (2017)

\bibitem{peng2018extreme}
Peng, B., Tan, W., Li, Z., Zhang, S., Xie, D., Pu, S.: Extreme network
  compression via filter group approximation. In: Proc. ECCV. pp. 300--316
  (2018)

\bibitem{pham2018efficient}
Pham, H., Guan, M., Zoph, B., Le, Q., Dean, J.: Efficient neural architecture
  search via parameters sharing. In: Proc. ICML. pp. 4095--4104 (2018)

\bibitem{real2019regularized}
Real, E., Aggarwal, A., Huang, Y., Le, Q.V.: Regularized evolution for image
  classifier architecture search. In: Proc. AAAI. vol.~33, pp. 4780--4789
  (2019)

\bibitem{ronneberger2015unet}
Ronneberger, O., Fischer, P., Brox, T.: U-{N}et: Convolutional networks for
  biomedical image segmentation. In: Proc. MICCAI. pp. 234--241. Springer
  (2015)

\bibitem{sandler2018mobilenetv2}
Sandler, M., Howard, A., Zhu, M., Zhmoginov, A., Chen, L.C.: Mobile{N}et{V}2:
  Inverted residuals and linear bottlenecks. In: Proc. CVPR. pp. 4510--4520
  (2018)

\bibitem{son2018clustering}
Son, S., Nah, S., Mu~Lee, K.: Clustering convolutional kernels to compress deep
  neural networks. In: Proc. ECCV. pp. 216--232 (2018)

\bibitem{sutton2018reinforcement}
Sutton, R.S., Barto, A.G.: Reinforcement learning: An introduction. MIT press
  (2018)

\bibitem{torfi2019gasl}
Torfi, A., Shirvani, R.A., Soleymani, S., Nasrabadi, N.M.: {GASL}: Guided
  attention for sparsity learning in deep neural networks. arXiv preprint
  arXiv:1901.01939  (2019)

\bibitem{wang2017factorized}
Wang, M., Liu, B., Foroosh, H.: Factorized convolutional neural networks. In:
  Proc. ICCVW. pp. 545--553 (2017)

\bibitem{wen2016learning}
Wen, W., Wu, C., Wang, Y., Chen, Y., Li, H.: Learning structured sparsity in
  deep neural networks. In: Proc. NeurIPS. pp. 2074--2082 (2016)

\bibitem{ye2018rethinking}
Ye, J., Lu, X., Lin, Z., Wang, J.Z.: Rethinking the
  smaller-norm-less-informative assumption in channel pruning of convolution
  layers. In: Proc. ICLR (2018)

\bibitem{yoon2017combined}
Yoon, J., Hwang, S.J.: Combined group and exclusive sparsity for deep neural
  networks. In: Proc. ICML. pp. 3958--3966. JMLR. org (2017)

\bibitem{yu2018nisp}
Yu, R., Li, A., Chen, C.F., Lai, J.H., Morariu, V.I., Han, X., Gao, M., Lin,
  C.Y., Davis, L.S.: {NISP}: Pruning networks using neuron importance score
  propagation. In: Proc. CVPR. pp. 9194--9203 (2018)

\bibitem{zeyde2010single}
Zeyde, R., Elad, M., Protter, M.: On single image scale-up using
  sparse-representations. In: International Conference on Curves and Surfaces.
  pp. 711--730. Springer (2010)

\bibitem{zhang2017beyond}
Zhang, K., Zuo, W., Chen, Y., Meng, D., Zhang, L.: Beyond a {G}aussian
  denoiser: residual learning of deep {CNN} for image denoising. IEEE TIP
  \textbf{26}(7),  3142--3155 (2017)

\bibitem{zhao2019variational}
Zhao, C., Ni, B., Zhang, J., Zhao, Q., Zhang, W., Tian, Q.: Variational
  convolutional neural network pruning. In: Proc. CVPR. pp. 2780--2789 (2019)

\bibitem{zoph2017nas}
Zoph, B., Le, Q.V.: Neural architecture search with reinforcement learning. In:
  Proc. ICLR (2017)

\bibitem{zoph2018nas}
Zoph, B., Vasudevan, V., Shlens, J., Le, Q.V.: Learning transferable
  architectures for scalable image recognition. In: Proc. CVPR (June 2018)

\end{thebibliography}

\appendix
\clearpage
\title{Supplementary Material for \\ ``DHP: Differentiable Meta Pruning via HyperNetworks''} % Replace with your title
% % \author{Yawei Li$^{1}$, Shuhang Gu$^{1,2}$, Kai Zhang$^1$, Luc Van Gool$^{1}$, Radu Timofte$^1$}
% % \institute{$^1$Computer Vision Lab, ETH Z\"urich, $^2$The University of Sydney, $^3$KU Leuven}

\titlerunning{DHP: Differentiable Meta Pruning via HyperNetworks} 
\authorrunning{Yawei Li, Shuhang Gu, Kai Zhang, Luc Van Gool, Radu Timofte} 
\author{Yawei Li$^{1}$, Shuhang Gu$^{1,2}$, Kai Zhang$^1$, Luc Van Gool$^{1,3}$, Radu Timofte$^1$}
\institute{$^1$Computer Vision Lab, ETH Z\"urich, $^2$The University of Sydney, $^3$KU Leuven \\
\email{\{yawei.li, kai.zhang, vangool, radu.timofte\}@vision.ee.ethz.ch} \\ \email{shuhanggu@gmail.com}}

\maketitle

\renewcommand\thefigure{S\arabic{figure}}    
\setcounter{figure}{0}    
\renewcommand\thetable{S\arabic{table}}    
\setcounter{table}{0}
\renewcommand{\theequation}{S\arabic{equation}}
\setcounter{equation}{0}

In this supplementary material, we first detail the training and testing protocol of the networks for different tasks in Sec.~\ref{sec:supp_protocol}. Then the latent vector sharing strategy is shown in Sec.~\ref{sec:supp_latent_vector_sharing}. More results are shown in Sec.~\ref{sec:supp_more_results}.

\section{Training and Testing Protocol}
\label{sec:supp_protocol}

As explained in the main paper, the proposed DHP method does not rely on the pretrained model. 
Thus, all of the networks are trained and pruned from scratch. 
The hypernetworks are first randomly initialized. Proximal gradient is used to sparsify the latent vectors. 
When the difference between the target and the actual FLOPs compression ratio is below 2\%, the pruning procedure stops. 
Then the pruned latent vectors as well as the pruned outputs of the hypernetworks are derived. 
After that, the outputs of hypernetworks are used as the weight parameters of the backbone network and updated by SGD or Adam algorithm directly. 
After the pruning procedure, the hypernetworks are removed. 
The training continues and the training protocol are the same as that utilized for training the original network. 
The number of pruning epochs is much smaller than that used for training the original network.
The following of the section describes the training protocols of different tasks. 

\subsection{Image Classification}
\label{subsec:supp_image_classification}

\subsubsection{CIFAR10}
\label{subsubsec:supp_cifar10}

CIFAR10~\cite{krizhevsky2009learning} contains 10 different classes. 
The training and testing subsets contain 50,000 and 10,000 images with resolution $32 \times 32$, respectively. 
As done by prior works~\cite{he2016deep,huang2017densely}, we normalize all images using channel-wise mean and standard deviation of the the training set. 
Standard data augmentation is also applied. 
The networks are trained for 300 epochs with SGD optimizer and an initial learning rate of 0.1. 
The learning rate is decayed by 10 after 50\% and 75\% of the epochs. 
The momentum of SGD is 0.9. Weight decay factor is set to 0.0001. The batch size is 64.

\subsubsection{Tiny-ImageNet}
\label{subsubsec:supp_tiny_imagenet}

For image classification, the pruning method is also compared on Tiny-Imagenet. 
It has 200 classes. Each class has 500 training images and 50 validation images. 
The resolution of the images is $64 \times 64$. 
The images are normalized with channel-wise mean and standard deviation. 
Horizontal flip is used to augment the dataset. 
The networks are trained for 220 epochs with SGD. 
The initial learning rate is 0.1. The learning rate is decayed by a factor of 10 at Epoch 200, Epoch 205, Epoch 210, and Epoch 215. 
The momentum of SGD is 0.9. Weight decay factor is set to 0.0001. The batch size is 64.

\subsection{Super-Resolution}
\label{subsec:supp_super_resolution}

\subsubsection{Training protocol}
\label{subsubsec:supp_protocol}

For image super-resolution, the networks are trained on DIV2K~\cite{Agustsson_2017_CVPR_Workshops} dataset. 
It contains 800 training images, 100 validation images, and 100 test images. 
Image patches are extracted from the training images. 
For EDSR, the patch size of the low-resolution input patch is $48 \times 48$ while for SRResNet the patch size is $24 \times 24$. The batch size is 16. 
The networks are optimized with Adam optimizer. 
The default hyper-parameter is used for Adam optimizer. 
The weight decay factor is 0.0001. The networks are trained for 300 epochs. 
The learning rate starts from 0.0001 and decays by 10 after 200 epochs.
The networks are tested on Set5~\cite{bevilacqua2012low}, Set14~\cite{zeyde2010single}, B100~\cite{MartinFTM01}, Urban100~\cite{Huang-CVPR-2015}, and DIV2K validation set. 

\subsubsection{Simplified EDSR architecture}
\label{subsubsec:supp_simplified_edsr}

In order to speed up the training of EDSR, a simplified version of EDSR is adopted.
The original EDSR contains 32 residual blocks and each convolutional layer in the residual blocks has 256 channels. 
The simplified version has 8 residual blocks and each has two convolutional layers with 128 channels.

\subsection{Denoising}
\label{subsec:supp_denosing}

For image denoising, the networks were trained on the gray version of DIV2K dataset and tested on BSD68 and DIV2K validation set. 
As done for image super-resolution, image patches are extracted from the training images. 
For DnCNN, the patch size of the input image is $64 \times 64$ and the batch size is 64. 
For UNet, the patch size is $128 \times 128$ and the batch size 16. 
Gaussian noise is added to degrade the input patches on the fly with noise level $\sigma = 70$. 
Adam optimizer is used to train the network. The weight decay factor is 0.0001. 
The networks are trained for 60 epochs and each epoch contains 10,000 iterations. 
So in total, the training continues for 600k iterations. 
The learning rate starts with 0.0001 and decays by 10 at Epoch 40.

\section{Latent Vector Sharing Strategy for Different Networks}
\label{sec:supp_latent_vector_sharing}

\subsection{Basic criteria}
\label{subsec:supp_basic_criteria}

To construct the hypernetworks, all of convolutional layers including standard convolution, depth-wise convolution, point-wise convolution, group convolution, and transposed convolution are attached a latent vector. 
The latent vectors act as the handle for network pruning. 
Thus, by dealing with only the latent vector, we can control how the convolutional layers are pruned.
But there could be complicated cases in the modern network architecture where the latent vectors have to be shared among different layers. 
Thus, during the development of the algorithm, we summarize some basic rules for latent vector sharing. 
In the following, we first describe the general rules for latent vectors and then detail the specific rules for special network blocks.

\begin{enumerate}
    \item[I] Every convolutional layer is attached a latent vector. 
    \item[II] The channel that the latent vector controls and the dimension of the latent vector vary with the types of convolutional layers. 
    \begin{enumerate}
        \item For standard convolution, point-wise convolution and transposed convolution, the latent vector controls the output channel of the layer and the dimension of the latent vector is the same as the number of output channels.
        \item For depth-wise convolution and group convolution, the latent vector controls the input channels per group. The dimension of the latent vector is the same as the number of input channels per group. That is, the latent vector of depth-wise convolution contains only one element. 
    \end{enumerate}
    \item[III] The latent vectors are shared among consecutive layers. This is because the output and input channels of consecutive layers are correlated. Thus, the hypernetworks receive the latent vectors of the previous layer and the current layer as input.
    \item[IV] Not every latent vector needs to be sparsified. The latent vectors free from sparsification are list as follows.
    \begin{enumerate}
        \item The latent vector that controls the input channel of the first convolutional layer. This latent vector has the same dimension with the input image channels, \eg 3 for RGB images and 1 for gray images. Of course, the input images do not need to be pruned.
        \item The latent vector attached to depth-wise convolution and group convolution. This latent vector controls the input channels per group. To compress depth-wise and group convolution, the number of groups is reduced, which is controlled by the latent vectors of the previous layer. 
    \end{enumerate}

\end{enumerate}

\subsection{Residual block}
\label{subsec:supp_residual_block}

The residual networks including ResNet, SRResNet, and EDSR are constructed by stacking a number of residual blocks. 
Depending on the dimension of the feature maps, the residual networks contain several stages with progressively reducing feature map dimension and increasing number of feature maps.
(Note that the feature map dimension of EDSR and SRResNet does not change for all of the residual blocks. So there is only one stage for those networks.) 
For the residual blocks within the same stage, their output channels are correlated due to the existence of the skip connections. 
In order to prune the second convolution of the residual blocks within the same stage, we use a shared latent vector for them. 
Thus, by only dealing with this shared latent vector, all of the second convolutions of the residual blocks can be pruned together. 
Please refer to Table~\ref{tbl:supp_sharing_vs_nonsharing} for the ablation study on latent vector sharing and non-sharing strategies. 

\subsection{Dense block}
\label{subsec:supp_dense_block}

Similar to residual networks, DenseNet also contains several stages with different feature map configurations. 
But different from residual networks, each dense block concatenates its input and output to form the final output of the block. 
As a result, each dense block receives as input the outputs of all of the previous dense blocks within the same stage. 
Thus, the hypernetwork of a dense block also has to receive the latent vectors of the corresponding dense blocks as input.  

\subsection{Inverted residual block}
\label{subsec:supp_inverted_residual_block}

The inverted residual blocks are just a special case of residual blocks. 
So how the latent vectors are shared across different blocks is the same with the normal residual blocks. 
Here we specifically address the sharing strategy within the block due to the existence of depth-wise convolution. 
The inverted residual block has the architecture of ``point-wise conv + depth-wise conv + point-wise conv''. 
As explained earlier, the latent vector of depth-wise convolution controls the input channels per group. 
Thus, the latent vector of the first point-wise convolution controls not only its output channels but also the input channels of the depth-wise convolution and the input channels of the second point-wise convolution. Thus, this latent vector has to be passed to the hypernetworks of the those convolutional layers. 

\subsection{Upsampler of super-resolution networks}
\label{subsec:supp_upsampler}

The image super-resolution networks are attached with upsampler blocks at the tail of the networks to increase the spatial resolution of the feature map. 
For the scaling factor of $\times 4$, two upsamplers are attached and each doubles the spatial resolution. 
Each of the upsampler block contains a standard convolutional layer that increases the number of feature maps by a factor of 4 and a pixel shuffler that shuffles every 4 consecutive feature maps into the spatial dimension. 
Thus, the output channel of the convolutional layer in the upsmapler is correlated to its input channel. 
If one input channel is pruned, then four corresponding consecutive output channels should also be pruned. 
To achieve this control of pruning, a common latent vector is used for the input and output channels.
The dimension of this latent vector is the same with the input channel size. 
This vector is repeated and interleaved to form the one controlling the output channel. 

\begin{table}[!hbt]
    \caption{Ablation study on ResNet56 for CIFAR10 image classification: exploring latent vector sharing strategy among correlated convolutional layers. ``Share'' denotes whether the latent vector sharing strategy described in Subsec.~\ref{subsec:supp_residual_block} is adopted. The $\ell_{2,1}$ regularizer means that when the latent vectors are not shared among the correlated layers, the group sparsity regularizer $\ell_{2,1}$ is enforced on their latent vectors. Otherwise, the normal $\ell_1$ sparsity regularizer is used. $\lambda$ and $\tau$ are the regularization factor and mask threshold introduced in the main paper. As shown in this table, the latent vector sharing strategy consistently outperforms the non-sharing counterparts}
    \label{tbl:supp_sharing_vs_nonsharing}
    \scriptsize
    \begin{center}
        \begin{tabular}{c|c|c|c|c|c|c|c}
            \toprule
            \multirow{2}{*}{Share} & \multirow{2}{*}{Regularizer} & \multirow{2}{*}{$\lambda$} & \multirow{2}{*}{$\tau$} & Target & Actual & Actual & Top-1 \\
            & & & & FLOPs Ratio (\%) & FLOPs Ratio (\%) & Parameter  Ratio (\%) & Error (\%) \\ \midrule
            Yes & $\ell_{1}$	& $2^{-4}$	& $5^{-3}$ & 	38 &	39.96 &	52.49 &	7.41 \\
            No	& $\ell_{1}$	& $2^{-4}$	& $5^{-3}$ &	38 &	39.75 &	55.43 &	7.32\\
            No	& $\ell_{2,1}$	& $2^{-4}$	& $5^{-3}$ &	38 &	39.40 &	54.24 &	7.91\\ \hline
            Yes & $\ell_{1}$	& $3^{-4}$	& $5^{-3}$ &	38 &	39.60 &	49.00 &	6.86\\
            No	& $\ell_{1}$	& $3^{-4}$	& $5^{-3}$ &	38 &	39.30 &	58.35 &	7.98\\
            No	& $\ell_{2,1}$	& $3^{-4}$	& $5^{-3}$ &	38 &	39.54 &	54.04 &	7.03\\ \midrule
            Yes & $\ell_{1}$	& $2^{-4}$	& $5^{-3}$ &	50 &	51.27 &	56.84 &	7.13\\
            No	& $\ell_{1}$	& $2^{-4}$	& $5^{-3}$ &	50 &	51.44 &	65.47 &	6.85\\
            No	& $\ell_{2,1}$	& $2^{-4}$	& $5^{-3}$ &	50 &	50.96 &	64.05 &	6.85\\ \hline
            Yes & $\ell_{1}$	& $3^{-4}$	& $5^{-3}$ &	50 &	51.68 &	57.74 &	6.52\\
            No	& $\ell_{1}$	& $3^{-4}$	& $5^{-3}$ &	50 &	50.23 &	62.83 &	7.11\\
            No	& $\ell_{2,1}$	& $3^{-4}$	& $5^{-3}$ &	50 &	50.18 &	59.15 &	6.74\\ \bottomrule
        \end{tabular}
    \end{center}
    % More results in the supplementary.}
\end{table}

\section{More Results}
\label{sec:supp_more_results}

The ablation study on the latent vector sharing strategy is shown in Table~\ref{tbl:supp_sharing_vs_nonsharing}. 
As shown in the table, the latent vector sharing strategy outperforms the non-sharing strategy consistently except for case $\lambda = 2^{-4}$, $\tau = 5^{-3}$ and 50\% target FLOPs compression ratio. 
The inconsistency is largely due to the gap between the actual parameter compression ratio of different strategies. 
Due to this fact, various latent vector sharing rules are developed for easier and better automatic network pruning.

The main paper utilizes $\ell_1$ regularizer to sparsify the the latent vectors. 
We tried to replace $\ell_1$ regularizer with $\ell_2$ regularizer and kept the same pruning strategy. 
The experiments are conducted on ResNet. 
The comparison results are shown in Table~\ref{tbl:supp_comparison_l1_l2}. 
Compared with $\ell_1$ regularizer, slightly worse results can be observed for $\ell_2$ regularizer. 
For ResNet-110 and ResNet-164, $\ell_2$ regularizer leads to results comparable with $\ell_1$ regularizer but at a larger parameter budget. 
When the regularizer is changed to $\ell_2$, the proximal operator becomes
\begin{equation}
    \mathbf{z}{[k+1]} = \left( 1 - \frac{\lambda\mu}{\mathrm{max} \left\{\|\mathbf{z}{[k+\Delta]}\|,\lambda\mu \right\}} \right)\mathbf{z}{[k+\Delta]}.
    \label{eqn:supp_proximal_operator_l2}
\end{equation}
By the formulation and experiments, $\ell_1$ norm leads to faster convergence. 
To have a reasonable convergence speed, the $\lambda$ used for $\ell_2$ regularizer is 20 times larger than that for $\ell_1$ regularizer.

\begin{table}[!hbt]
    \caption{Comparison between $\ell_1$ norm and $\ell_2$ norm regularization. The experiments are done on ResNet}
    \label{tbl:supp_comparison_l1_l2}
    \scriptsize
    \begin{center}
        \begin{tabular}{c|c|c|c|c}
            \toprule
            Layer	&Regularizer	&Top1 Error	(\%) &FLOPs (\%)	&Params (\%)\\ \hline
            20	& $\ell_1$	&8.46	&51.8	&56.13 \\ 
                & $\ell_2$	&8.66	&51.59	&54.19 \\ \hline
            110	& $\ell_1$	&5.73	&51.62	&54.13 \\ 
                & $\ell_2$	&5.77	&51.37	&72.37 \\ \hline
            164	& $\ell_1$	&5.22	&51.67	&50.97 \\ 
                & $\ell_2$	&5.18	&50.87	&60.66 \\
            \bottomrule
        \end{tabular}
    \end{center}
\end{table}

\begin{figure}[t]
\centering     %%% not \center
\subfigure[FLOP compression ratio.]{\label{fig:supp_resnet_164_110_dhp_flops}\includegraphics[width=0.45\textwidth]{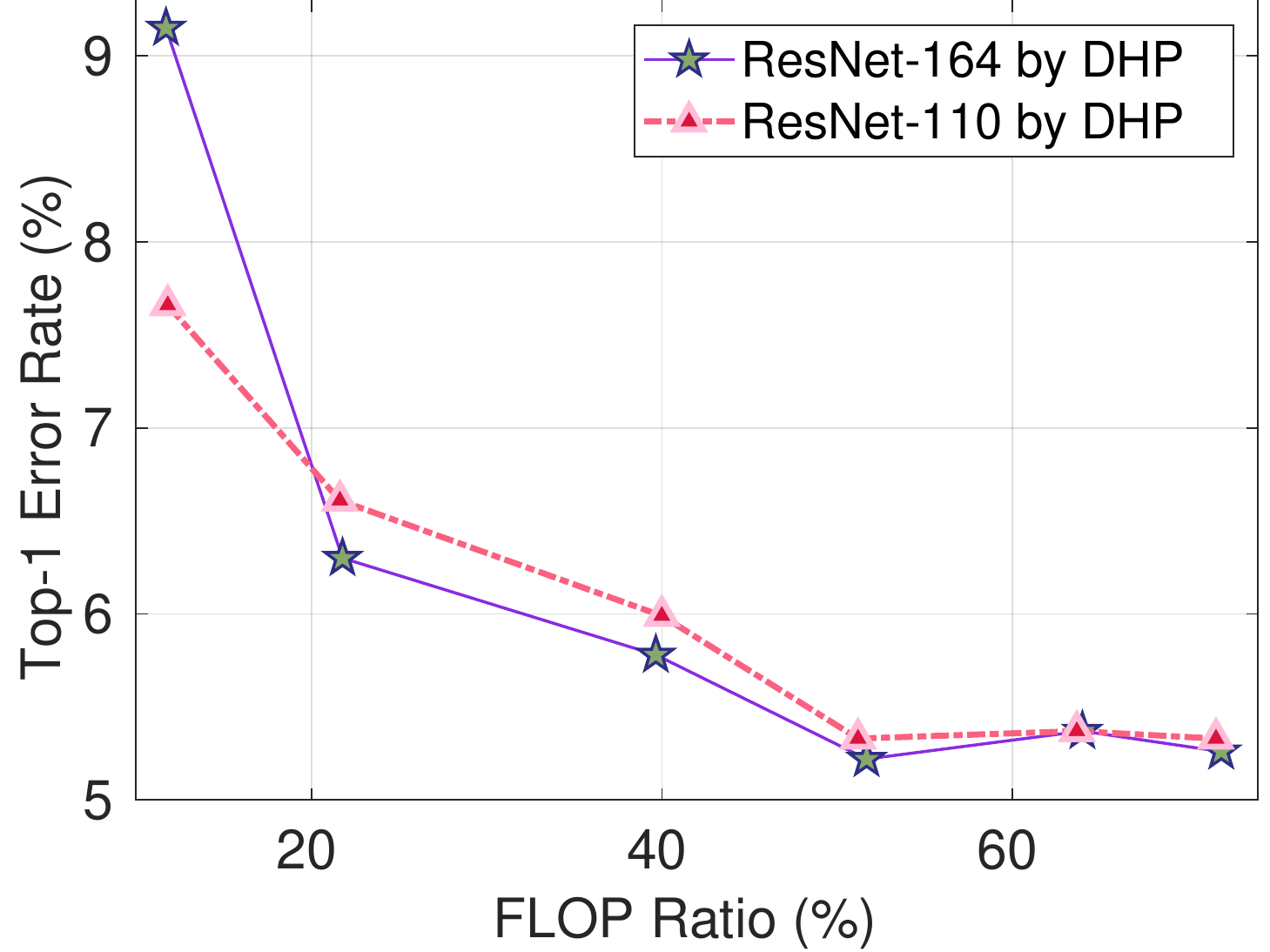}}
\subfigure[Parameter compression ratio.]{\label{fig:supp_resnet_164_110_dhp_params}\includegraphics[width=0.45\textwidth]{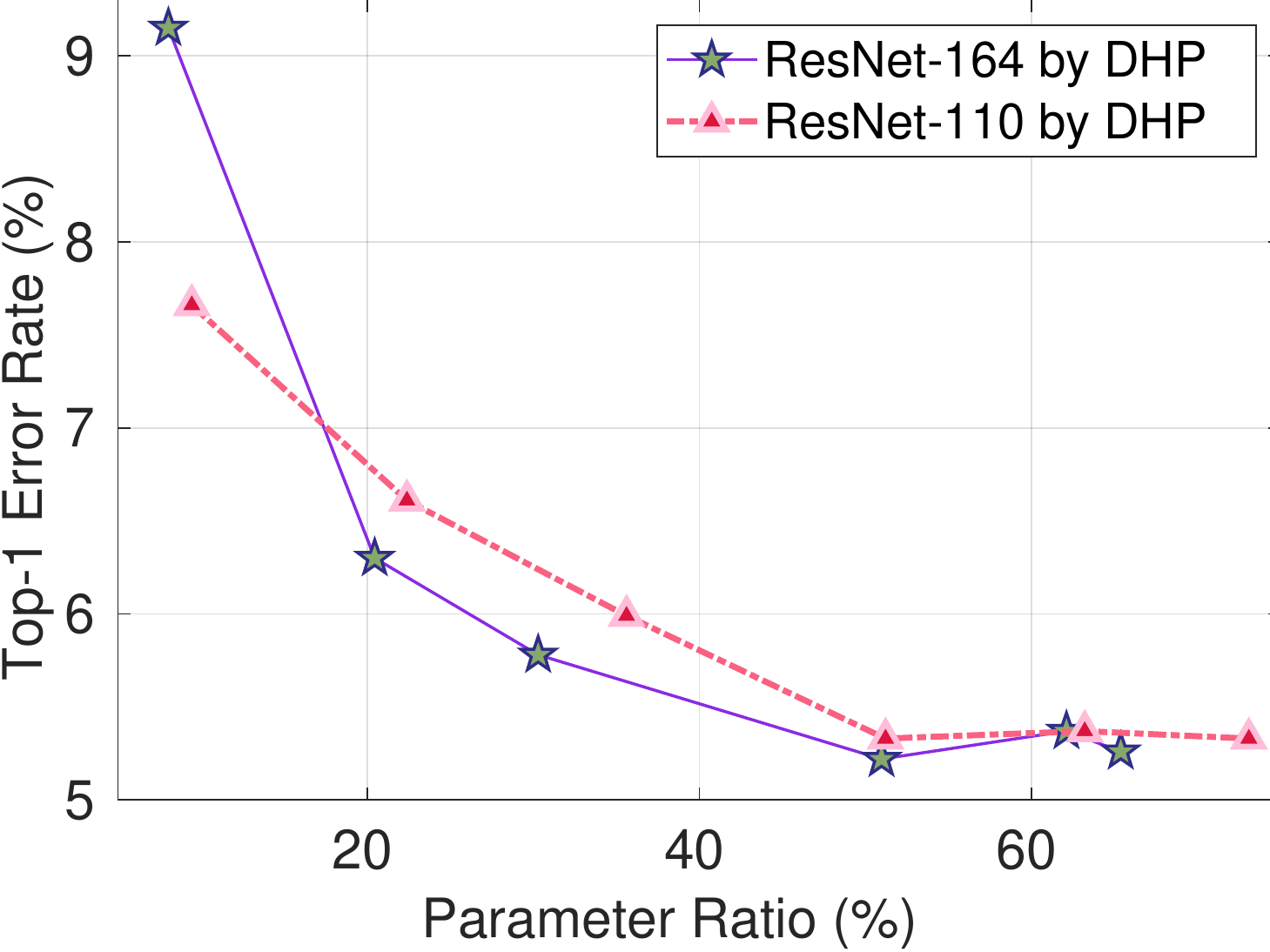}}
\caption{Top-1 error \vs FLOP and parameter compression ratio on ResNet-164 and ResNet-110}
\label{fig:supp_resnet110_resnet164}
\end{figure}

More results on ResNet-110 and ResNet-164 are shown in Fig.~\ref{fig:supp_resnet110_resnet164}. 
When the compression ratio is not too severe (above 50\%), the accuracy does not drop too much. 
The extreme compression prunes about 90\% FLOPs and parameters of the original network. 
For ResNet-164, the extreme compression only keeps 8.04\% parameters. 
Thus, the drop in the accuracy is reasonable. 

\end{document}